\UseRawInputEncoding
\documentclass[journal]{IEEEtran}
\pdfoutput=1
\usepackage{url}
\usepackage{hyperref}
\usepackage{breakurl}

\usepackage{amsmath,amsfonts}
\usepackage{algorithmic}
\usepackage{array}
\usepackage[caption=false,font=normalsize,labelfont=sf,textfont=sf]{subfig}
\usepackage{textcomp}
\usepackage{stfloats}
\usepackage{url}
\usepackage{verbatim}
\usepackage{graphicx}
\usepackage{cite}
\usepackage{balance}
\usepackage{amssymb}
\usepackage{amsthm}
\usepackage{amsmath}
\usepackage{multirow}
\usepackage{graphicx}
\usepackage{tabularx}
\usepackage[linesnumbered, ruled]{algorithm2e}
\SetKwRepeat{Do}{do}{while}%
\begin{document}

\title{Generalized Time Warping Invariant Dictionary Learning for Time Series Classification and Clustering}

\author{Ruiyu Xu, Chao Wang, Yongxiang Li, Jianguo Wu$^*$
        % <-this % stops a space
\thanks{R. Xu and J. Wu are with the Department of Industrial Engineering and Management, Peking University, Beijing, China.}% <-this % stops a space
\thanks{C. Wang is with the Department of Industrial and Systems Engineering, University of Iowa, Iowa City, IA.}% <-this % stops a space
\thanks{Y. Li is with the Department of Industrial Engineering and Management, Shanghai Jiao Tong University, Shanghai, China.}
\thanks{*Corresponding author: Jianguo Wu (j.wu@pku.edu.cn)}
\thanks{This work has been submitted to the IEEE for possible publication. Copyright may be transferred without notice, after which this version may no longer be accessible.}}



\maketitle

\begin{abstract}
Dictionary learning is an effective tool for pattern recognition and classification of time series data. Among various dictionary learning techniques, the dynamic time warping (DTW) is commonly used for dealing with temporal delays, scaling, transformation, and many other kinds of temporal misalignments issues. However, the DTW suffers overfitting or information loss due to its discrete nature in aligning time series data. To address this issue, we propose a generalized time warping invariant dictionary learning algorithm in this paper. Our approach features a generalized time warping operator, which consists of linear combinations of continuous basis functions for facilitating continuous temporal warping. The integration of the proposed operator and the dictionary learning is formulated as an optimization problem, where the block coordinate descent method is employed to jointly optimize warping paths, dictionaries, and sparseness coefficients. The optimized results are then used as hyperspace distance measures to feed classification and clustering algorithms. The superiority of the proposed method in terms of dictionary learning, classification, and clustering is validated through ten sets of public datasets in comparing with various benchmark methods.
\end{abstract}


\section{Introduction}
\label{sec1}
Dictionary learning represents input time series data as the combination of a few basis functions, known as atoms. These atoms contain critical temporal features of the original time series data, thus are defined as the dictionary for the data. The dictionary learning has been successfully applied in various tasks such as feature extraction \cite{liu2016dictionary}, reconstruction \cite{caballero2014dictionary}, denoising \cite{dong2011sparsity}, compressed sensing \cite{chen2014compressed}, classification \cite{yang2014sparse, ramirez2010classification}, and clustering \cite{dong2011sparsity, ramirez2010classification}. However, traditional dictionary learning assumes that input data is well-aligned with atoms, which is often violated in practice. For example, biological processes exhibit variability in rate and progress across organisms, strains, individuals, and conditions, which pose difficulties for identifying disease progression among affected individuals \cite{aach2001aligning}. Such misalignment also makes it challenging for traditional methods to reveal latent dictionaries and achieve accurate prediction in subsequent tasks.

In recent decades, various techniques have been proposed to address the misalignment in dictionary learning. They can be broadly classified into three categories: kernel approaches, shapelet learning approaches, and dynamic time warping (DTW, \cite{berndt1994using}) based approaches.

The Kernel approaches \cite{scholkopf2005kernel, van2012kernel, yin2012kernel, chen2015kernel, zhang2011kernel} map time series data into an implicit reproducing kernel Hilbert space, and dictionaries and atoms are learned in this implicit space. Although kernel approaches are flexible in dealing with various kinds of misalignments, their operations and learning in implicit space raises practical concerns in interpreting the results \cite{deng2020invariant}. Moreover, the performance of kernel approaches highly depends on the selection of kernels and their parameters, which impairs their performance and robustness in subsequent tasks of classification and clustering \cite{deng2020invariant}.

The shapelet approaches \cite{lewicki1998coding, mailhe2008shift, yang2017fault, zheng2016efficient, li2021dictionary,deng2018robust} represent the original time series data as various local atoms, known as shapelets, which can be used to extract local, phase-independent similarity in shape. Although these methods are capable of learning representative local patterns, there lacks an efficient and interpretable way for integrating local shapelets into a complete time series data. This limits the shapelet methods in providing a comprehensive analysis of global features. Moreover, shapelet approaches require a large training dataset to capture sufficient shapelet candidates, which can be computationally expensive in terms of time and resources.

To provide a more interpretive analysis of global patterns, the third category of approaches \cite{deng2020invariant,yazdi2018time,yazdi2019time} introduces a kind of warping operator to align the input time series with a dictionary based on DTW alignment or its variants. In this framework, the sparse coding, dictionary, and warping operators are jointly optimized. However, the effectiveness of these methods is heavily influenced by the DTW alignment performance, which could be compromised due to multiple factors. Firstly, DTW is sensitive to noise, as small fluctuations in the time series can result in significant changes in the alignment. It may produce inaccurate warping outcomes in the presence of noise. Secondly, DTW may yield counterintuitive and suboptimal alignments due to its discrete nature. This could result in a single point from one time series being mapped to a section of another time series. Such alignments would lead to overfitting and information loss about the subtle details. Thirdly, DTW assumes the start and end points of the two time series must be aligned, which is known as the boundary condition. This assumption hinders the application of DTW in aligning partial data among different time series, which is commonly encountered in signal extraction/enhancement.

To address these issues in DTW based dictionary learning, this paper proposes a novel method called Generalized Time Warping Invariant Dictionary Learning (GTWIDL) that relaxes warping boundaries, expands applications, facilitates interpretability, and improves performance of dictionary learning. More specifically, the proposed method parameterizes the time warping operators as a linear combination of a few monotonic functions, which achieves more accurate alignments and relaxes the boundary conditions. Our model also features a $l_1$ Lasso penalty to ensure sparseness of parameters, which is especially time-efficient when dealing with multi-dimensional time series data. An iterative optimization algorithm is further developed to sequentially update the sparseness coefficients, dictionary, and warping operators. Case studies show that this optimization algorithm consistently demonstrates high accuracy and convergence for various datasets. Based on the optimized results, time series classification and clustering algorithms are further developed to take advantages of the improved dictionary performance to achieve superior accuracy in corresponding tasks.

The contributions of the paper are as follows:
\begin{itemize}
	\item We propose a novel framework for time warping invariant dictionary learning. By formulating the warping operators as a continuous and unconstrained warping path, our approach addresses the limitations of traditional DTW-based methods and facilitates joint alignment of multi-dimensional time series data.
	
	\item We develop an efficient optimization algorithm to jointly update the sparseness coefficients, dictionary, and warping operators.
	
	\item Built upon the proposed dictionary learning, interpretable classification and clustering algorithms are developed to extract representative atoms for each class, which improves the explanatory power of the generated models.
	
	\item The proposed methods are evaluated on various public datasets consisting of both one-dimensional and multi-dimensional time series data. The experimental analysis indicates that the proposed algorithm surpasses the performance of traditional dictionary learning methods and existing time warping invariant techniques.
\end{itemize}

The remainder of this paper is organized as follows. Section \ref{sec2} provides relevant background information and a review of related works. In Section \ref{sec3}, we present our proposed generalized time warping invariant dictionary learning framework, optimization algorithm and related analysis. Time series classification and clustering algorithms are further developed in Section \ref{sec4}. We evaluate the proposed methods on several representative datasets to demonstrate their effectiveness in Section \ref{sec5}. Finally, Section \ref{sec6} provides a brief conclusion of this study.

\section{Related work}
\label{sec2}
In this section, we present a brief overview of the existing literature concerning temporal alignment and time warping invariant classification and clustering methods.

\subsection{Temporal alignment}
\label{sec21}
Misaligned data refers to data that are not appropriately synchronized along the time axis. This phenomenon may occur when data are collected from multiple sources or when data are recorded at different times or locations. Temporal alignment aims to synchronize these misaligned data streams or time series so that features and patterns can be extracted more efficiently. Temporal alignment has garnered significant attention from researchers, where the Dynamic Time Warping (DTW, \cite{berndt1994using}) is one of the most well-known algorithms. By locally stretching and compressing time series, DTW provides a pairwise match between two time series that minimizes their Euclidean difference after alignment. Specifically, DTW is defined as follows.

Given two time series, $\boldsymbol{\rm x} = \left[x_1, x_2, \ldots, x_{n_x}\right]\in\mathbb{R}^{n_x}$ and $\boldsymbol{\rm y} = \left[y_1, y_2, \ldots, y_{n_y}\right]$ $\in\mathbb{R}^{n_y}$, DTW aligns $\boldsymbol{\rm x}$ and $\boldsymbol{\rm y}$ such that the sum of the distances between the aligned samples is minimized:
\begin{equation}
	\label{equ1}
	\mathop{\min}\limits_{\left\{\boldsymbol{p}^x,\boldsymbol{p}^y\right\}\in \boldsymbol{\Psi}} d_{DTW} \left( \boldsymbol{\rm x} , \boldsymbol{\rm y} \right) = \sum_{t=1}^{l} \left \| x_{p_t^x} - y_{p_t^y} \right \|_2^2,
\end{equation}
where $\boldsymbol{p}^x \in \left\{1:n_x\right\}^l$ and $\boldsymbol{p}^y \in \left\{1:n_y\right\}^l$ are the warping paths of time series $\boldsymbol{\rm x}$ and $\boldsymbol{\rm y}$ respectively, $l\ge \max{\left(n_x,n_y\right)}$ is the number of indices used to align the samples, and $l$ is automatically selected by the DTW algorithm. The $i$th frame in $\boldsymbol{\rm x}$, i.e., $x_i$, and the $j$th frame in $\boldsymbol{\rm y}$, i.e., $y_j$, are aligned if there exist $p_t^x = i$ and $p_t^y = j$ at some timestamp $t$. $\boldsymbol{\Psi}$ gives the constraints that warping paths $\boldsymbol{p}^x$ and $\boldsymbol{p}^y$ must satisfy, including the boundary, monotonicity and continuity constraints:
\begin{itemize}
	\item Boundary: $p_1^x = 1, p_1^y = 1, p_l^x = n_x$ and $p_l^y = n_y$.
	\item Monotonicity: $t_1>t_2 \Rightarrow p_{t_1}^x \ge p_{t_2}^x$ and $p_{t_1}^y \ge p_{t_2}^y$.
	\item Continuity: \\$\left[p_t^x , p_t^y \right] - \left[p_{t-1}^x , p_{t-1}^y \right] \in \left\{\left[1,1\right],\left[0,1\right],\left[1,0\right]\right\}$.
\end{itemize}
The DTW algorithm efficiently solves this optimization problem via a dynamic programming approach. The distance between the two time series in their $i$th and $j$th frames, i.e., $d_{ij}$, can be calculated using their current distance and the minimal value of their distance in their previous frames:
\begin{equation}
	d_{ij} =\mathop{\min}\limits_{\pi_{ij}^x,\pi_{ij}^y}  d_{i-\pi_{ij}^x,j-\pi_{ij}^y} +\left ( x_{i} - y_{j} \right )^2, d_{11} = \left ( x_{1} - y_{1} \right )^2,
\end{equation}
where $\boldsymbol{\pi}_{ij}=\left[\pi_{ij}^x,\pi_{ij}^y\right]$ stands for the optimal policy queue for the two time series in their $i$th and $j$th frames. Once the policy queue is known, the alignment steps can be recursively selected by backtracking, $\left[p_{t-1}^x , p_{t-1}^y \right] = \left[p_t^x , p_t^y \right] - \boldsymbol{\pi}_{p_t^x,p_t^y}, p_l^x = n_x$ and $p_l^y = n_y$. Warping matrices $\boldsymbol{W}_x\in \{0,1\}^{n_x \times l},\boldsymbol{W}_y\in \{0,1\}^{n_y \times l} $ are further built based on the warping path $p_{1:l}^x$ and $p_{1:l}^y$ to calculate $x_{p_t^x}$ and $y_{p_t^x}$. The element $\boldsymbol{W}_x^{(i,t)}$ equals 1 if $p_t^x=i$ and 0 else. Denote the aligned time series as $\boldsymbol{\rm x}_{\boldsymbol{p}^x}$ and $\boldsymbol{\rm y}_{\boldsymbol{p}^y}$. We have $\boldsymbol{\rm x}_{\boldsymbol{p}^x}=\boldsymbol{\rm x}\boldsymbol{W}_x$ and $\boldsymbol{\rm y}_{\boldsymbol{p}^y}=\boldsymbol{\rm y}\boldsymbol{W}_y$, then Equation (\ref{equ1}) turns to: 
\begin{equation}
	\mathop{\min}\limits_{\left\{\boldsymbol{p}^x,\boldsymbol{p}^y\right\}\in \boldsymbol{\Psi}} d_{DTW} \left( \boldsymbol{\rm x} , \boldsymbol{\rm y} \right) = \left \|\boldsymbol{\rm x}\boldsymbol{W}_x \left(\boldsymbol{p}^x \right)-\boldsymbol{\rm y}\boldsymbol{W}_y \left(\boldsymbol{p}^y \right)\right \|_2^2.
\end{equation}
The computational cost of DTW algorithm is $O(n_x n_y)$ in space and time \cite{salvador2007toward}.

A significant amount of research has been dedicated to accelerating the computation of DTW algorithm and improving the control of warping path routes. Global constraints on warping paths, such as the Sakoe-Chiba band and Itakura Parallelogram band \cite{rabiner1993fundamentals}, have been proposed to reduce the number of possible alignment routes. Multi-level searching approaches \cite{salvador2007toward,chu2002iterative} have also been developed to iteratively increase the precision of warping paths, resulting in a considerable speedup of up to three orders of magnitude compared to the classic DTW algorithm. Furthermore, feature extraction from raw data has been explored to provide a more in-depth understanding of alignment, including derivatives \cite{2002Derivative}, phase differences \cite{jeong2011weighted}, and local structures \cite{xie2010adaptive,zhao2018shapedtw}. Despite the progress in improving the DTW, these methods still belong to the discrete warping framework, leading to less accurate warping paths. In particular, a single frame of one time series may be assigned to many consecutive frames in the other time series, causing overfitting or information loss. To address this issue, Zhou et al. \cite{zhou2012generalized,zhou2015generalized} proposed a Generalized Canonical Time Warping (GCTW) method. The GCTW incorporates a linear combination of monotonic functions to represent the warping path and provides a more flexible and continuous temporal warping. GCTW also relaxes the boundary constraint in regular DTW.

\begin{figure}[tb]
	\centering
	\includegraphics[width=\columnwidth]{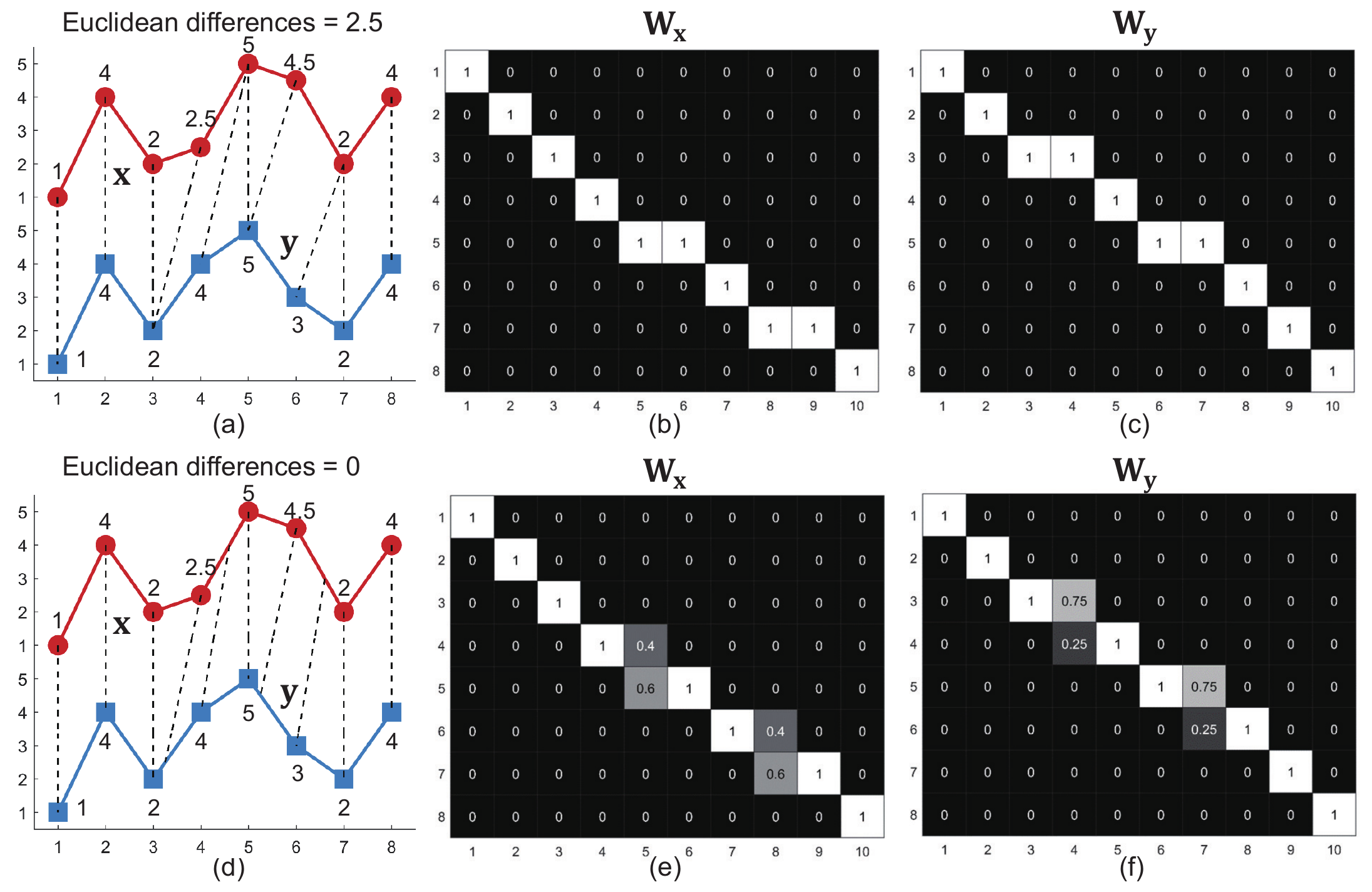}
	\caption{An example of aligning time series in discrete and continuous ways. (a) Two 1-D time series ($n_x=n_y=8$) and the discrete alignment between samples computed by DTW. (b-c) Discrete warping matrices for two time series. (d) Continuous alignment between samples. (e-f) Continuous warping matrices for two time series.}
	\label{example_DTW}
\end{figure}

Fig. \ref{example_DTW} shows an example of temporal alignment in both discrete and continuous manners. The discrete alignment is computed by DTW, which provides pairwise alignment between points of two time series presented in Fig. \ref{example_DTW}(a). The resulting alignment shows the occurrence of one-to-many alignments at frames 5 and 7 in time series $\boldsymbol{\rm x}$ and frame 3 and 5 in time series $\boldsymbol{\rm y}$. In Fig. \ref{example_DTW}(b-c), the discrete warping matrices for the two time series are depicted, and their elements are binary, determined by the warping paths. On the other hand, Fig. \ref{example_DTW}(d) shows the continuous alignment where single frames of one time series can be assigned to some unobserved frames using the interpolation approach. The continuous warping matrices in Fig. \ref{example_DTW}(e-f) present their elements as decimals ranging from 0 to 1, and the sum of each column is equal to 1. The use of continuous warping paths allows for more precise assignment with smaller Euclidean differences after alignment. As an illustration, let us consider the scenario of aligning frame 4 in $\boldsymbol{\rm x}$ with frame 3 in $\boldsymbol{\rm y}$. A discrete alignment yields a Euclidean difference of 0.25 between the two frames. In contrast, a continuous alignment achieved through interpolation of frame 3 and 4 in $\boldsymbol{\rm y}$ results in a complete alignment and a Euclidean difference of zero.

\subsection{Time warping invariant classification and clustering}
\label{TWICC}

The DTW has been widely implemented in time series classification and clustering for handling misaligned data. The DTW can be directly used as a distance measure, and they can also be combined with other techniques, e.g., dictionary learning, for various classification and clustering purposes, e.g., fast computation and multi-pattern recognition. We briefly review these techniques as follows.

In the context of classification problems, G{\'o}recki et al. \cite{gorecki2014non} proposed a fuzzy distance function based on derivatives and transforms. Jeong et al. \cite{jeong2011weighted} proposed a weighted derivative dynamic time warping (WDDTW) method, which weighs each point according to the phase difference between a test point and a reference point. These methods, along with benchmarks like the classic DTW method \cite{berndt1994using} and Derivative Dynamic Time Warping (DDTW, \cite{2002Derivative}), realize the classification of test samples by assigning to the class of their nearest neighbor in the 1-Nearest Neighbor (1NN) classifier. Besides the 1NN classifier, the DTW distances can also be fed as input features to other classifiers. Kate \cite{kate2016using}, for example, employed DTW to create similarity features that are then given to a Support Vector Machine (SVM) classifier. For clustering problems, Du et al. \cite{du2019dynamic} used DTW to calculate the similarity matrix, based on which the spectral clustering method was applied without training steps. Lohrer et al. \cite{lohrer2015building} built the similarity matrix via the FastDTW method, leading to a significant reduction of computation time compared to full DTW. The authors also employed spectral clustering to obtain clustering outcomes.

However, these DTW distance-based methods face challenges when classifying or clustering large datasets. For example, classification tasks require computing DTW distances between all training samples and test samples, consequently leading to high computation complexity. Additionally, DTW distance-based methods only consider distance information, disregarding warping information, which limits their ability in extracting data patterns. Similar issues also exist in clustering tasks. Therefore, some works have focused on the combination of the DTW technique and centroid-based classification and clustering methods. In the classification framework, Petitjean et al. \cite{petitjean2016faster} used DTW Barycenter Averaging (DBA) to calculate the centroids of classes used in Nearest Centroid Classifier (NCC). Soheily et al. \cite{soheily2016generalized} proposed a generalized k-means-based clustering method that employs weighted and kernel time warping to assign test samples to the nearest centroid. These methods significantly reduce computation complexity, and the learned centroids of all classes offer additional understanding and interpretation of classification criterion. In the clustering framework, similar approaches combining DBA technique and K-means method are implemented \cite{petitjean2011global}. Izakian et al. \cite{izakian2015fuzzy} further proposed a fuzzy K-Medoids clustering method, where K samples from the dataset are selected as cluster centers to overcome challenges in calculating the average of time series collections. The authors also combined fuzzy K-means and fuzzy K-Medoids methods to develop a hybrid clustering method, which exhibited improved clustering performance.

These centroid-based methods assume that there is only one centroid per class, failing to account for the presence of multiple patterns within a class. To fill this gap, dictionary learning methods have been utilized to incorporate multiple atoms for modeling the patterns. One such method was proposed by Yazdi et al. \cite{yazdi2018time}, which introduced a cosine maximization time warping operator (COSTW) and optimized warping paths, dictionaries, and sparse representing coefficients to train a classifier. This method is extended into a clustering framework by iterative updating of dictionaries and clustering results \cite{yazdi2019time}. Besides, Deng et al. \cite{deng2020invariant} employed an SVM classifier using representing coefficients on the global dictionary as features, and used an Efficiently Learning Shapelets (ELS, \cite{hou2016efficient}) feature selection model to improve classification accuracy. 

Nevertheless, as we mentioned in Section \ref{sec21} and demonstrated in Fig. \ref{example_DTW}, the DTW suffers intrinsic deficiency due to its discrete nature in aligning time series data. This deficiency in data alignment will be propagated along the modeling and analysis methods and finally impairs the performance of classification and clustering that use DTW aligned data as inputs. As a result, a novel time warping invariant classification and clustering method is needed to acquire more accurate alignment and warping paths and achieve improved performance in classification and clustering.

\section{Time warping invariant dictionary learning}
\label{sec3}
In this section, we present the formalization of the proposed time series dictionary learning that is invariant to time warping. Instead of using the DTW method, we adopt the concept of the GCTW method and describe warping paths in terms of linear combinations of predefined monotonic functions. Additionally, we propose an efficient optimization algorithm that alternatively updates dictionaries, warping paths, and sparseness coefficients to achieve optimal values. Finally, we provide an analysis of the proposed dictionary learning method, including a discussion of its computational complexity and its relationship to existing approaches.

\subsection{Problem statement}
Given a collection of $m$ one-dimensional time series, $\left\{\boldsymbol{\rm x}_i\right\}_{i=1}^{m}$, traditional dictionary learning methods seek a dictionary $\boldsymbol{D}$ with a set of $K$ time series atoms $\left\{\boldsymbol{d}_k\right\}_{k=1}^{K}$, and the sparse representing coefficients $\boldsymbol{\alpha}_i = \left[\alpha_{i1}; \ldots; \alpha_{iK}\right] \in \mathbb{R}^K$ for each $\boldsymbol{\rm x}_i = \left[x_1^i, x_2^i, \ldots, x_{n}^i\right] \in\mathbb{R}^{1 \times n}$. Such optimization problem could be solved by minimizing the mean squared error between the input and their reconstructions:
\begin{equation}
	\begin{gathered}
		\mathop{\min}\limits_{\left\{\boldsymbol{d}_k\right\}_{k=1}^{K},\left\{\boldsymbol{\alpha}_i\right\}_{i=1}^{m}}  \sum_{i=1}^{m}  \left\{\frac{1}{n}\left \| \boldsymbol{\rm x}_i - \sum_{k=1}^{K} \alpha_{ik}\boldsymbol{d}_k  \right \|_2^2 + \lambda \left \| \boldsymbol{\alpha}_i \right \|_1 \right\} ,\\
		\text{s.t.} \ \ \left\| \boldsymbol{d}_{k} \right\|_2=1 ,
	\end{gathered}
\end{equation}
where the $l_1$ penalty guarantees the sparsity of representing coefficients and $\lambda$ represents the penalty parameter.

To address data misalignment, we introduce warping matrices $\left\{\boldsymbol{W}_i \right\}_{i=1}^{m}$ into the optimization problem. These matrices facilitate time warping of each sample, i.e., $\boldsymbol{\rm x}_i$, such that it aligns well with the dictionary. Furthermore, we impose a constraint on the sparseness coefficients $\boldsymbol{\alpha}_{i}\ge\boldsymbol{0}$, which ensures the coherence of trends between samples and atoms. This constraint aims to reduce the feasible region of the dictionary. Otherwise, the variables $\left\{\tilde{\boldsymbol{\alpha}}_i\right\}_{i=1}^{m}=\left\{-\boldsymbol{\alpha}_i\right\}_{i=1}^{m}$ and $\left\{\tilde{\boldsymbol{d}}_k\right\}_{k=1}^{K}=\left\{-\boldsymbol{d}_k\right\}_{k=1}^{K}$ would produce the same results. As a result, the $l_1$ penalty is equivalent to a reduced linear function of $\boldsymbol{\alpha}_i$.
\begin{footnotesize}
\begin{equation}
	\begin{gathered}
		\mathop{\min}\limits_{\left\{\boldsymbol{d}_k\right\}_{k=1}^{K},\left\{\boldsymbol{\alpha}_i\right\}_{i=1}^{m}, \left\{\boldsymbol{W}_i \right\}_{i=1}^{m}}  \sum_{i=1}^{m}  \left\{\frac{1}{n_i}\left \| \boldsymbol{\rm x}_i - \sum_{k=1}^{K} \alpha_{ik}\boldsymbol{d}_k \boldsymbol{W}_i  \right \|_2^2 + \lambda \boldsymbol{1}' \boldsymbol{\alpha}_i \right\} ,\\
		\text{s.t.} \ \ \boldsymbol{\alpha}_{i}\ge\boldsymbol{0},\boldsymbol{1}_{n_D}'  \boldsymbol{W}_i = \boldsymbol{1}_{n_i}', \left\| \boldsymbol{d}_{k} \right\|_2=1 ,
	\end{gathered}
\end{equation}
\end{footnotesize}
where the length of each sample and the length of atoms may differ, $\boldsymbol{\rm X}_i \in\mathbb{R}^{1 \times n_i}, \boldsymbol{d}_k \in \mathbb{R}^{1 \times n_D}$, and the warping matrix $\boldsymbol{W}_i \in \mathbb{R}^{n_D \times n_i}$ is a presentation of the monotonous warping path. The predetermined length of atoms $n_D$ is selected as the average length of training samples in our research. Such a framework can be extended to a $p$-dimensional case.
\begin{footnotesize}
\begin{equation}
	\label{opt0}
	\begin{gathered}
		\mathop{\min}\limits_{\left\{\boldsymbol{D}_j\right\}_{j=1}^{p},\left\{\boldsymbol{\alpha}_i\right\}_{i=1}^{m}, \left\{\boldsymbol{W}_i \right\}_{i=1}^{m}}  \sum_{i=1}^{m}  \left\{\frac{1}{n_i}\sum_{j=1}^{p}\left \| \boldsymbol{\rm x}_{ij} -  \boldsymbol{\alpha}_{i}^{T}\boldsymbol{D}_{j} \boldsymbol{W}_i  \right \|_2^2 + \lambda \boldsymbol{1}' \boldsymbol{\alpha}_i \right\} ,\\
		\text{s.t.} \ \ \boldsymbol{\alpha}_{i}\ge\boldsymbol{0},\boldsymbol{1}_{n_D}'  \boldsymbol{W}_i = \boldsymbol{1}_{n_i}' , \left\| \boldsymbol{d}_{jk} \right\|_2=1,
	\end{gathered}
\end{equation}
\end{footnotesize}
where $\boldsymbol{\rm X}_i$ is a $p$-dimensional time series $\boldsymbol{\rm X}_i \in\mathbb{R}^{p \times n_i}$ and the data in $j$-th dimension is denoted as $\boldsymbol{\rm x}_{ij}$. Accordingly, $\boldsymbol{d}_{jk} \in \mathbb{R}^{1 \times n_D}$ is the $k$-th atom in $j$-th dimension and $\boldsymbol{D}_{j}=\left[\boldsymbol{d}_{j1};\ldots;\boldsymbol{d}_{jK}\right]\in\mathbb{R}^{K\times n_D}$ denotes the dictionary in $j$-th dimension. 

The proposed framework (\ref{opt0}) facilitates multi-pattern recognition for multi-dimensional time series data with $K$ atoms. In addition, compared with the existing dictionary learning approaches \cite{deng2020invariant,yazdi2018time,yazdi2019time} mentioned in Section \ref{TWICC}, it significantly reduces the number of variables to be optimized and effectively reduces the computational complexity by sharing the sparseness coefficients $\boldsymbol{\alpha}_i$ and the warping matrix $\boldsymbol{W}_i$ across all dimensions. Despite its advantages, the formulation in (\ref{opt0}) is still based on DTW, which suffers various issues as mentioned in Section \ref{TWICC} and demonstrated in Fig. \ref{example_DTW}.

In order to address the aforementioned weaknesses of DTW, we parameterize the warping path $\boldsymbol{p}_i$ as a linear combination of monotonic functions
\begin{equation}
	\boldsymbol{p}_i = \boldsymbol{Q}_i\boldsymbol{\beta}_i,
\end{equation}
where $\boldsymbol{Q}_i=\left[\boldsymbol{q}_{i1},\ldots,\boldsymbol{q}_{i\tilde{k}}\right]\in \mathbb{R}^{n_i\times \tilde{k}}$ is the basis set composed of $\tilde{k}$ pre-defined monotonically increasing functions and $\boldsymbol{\beta}_i\in \mathbb{R}^{\tilde{k}}, \boldsymbol{\beta}_i\ge0$ is the non-negative coefficients. The monotonicity property of warping path $\boldsymbol{p}_i$ is preserved due to the fact that any non-negative combination of monotonically increasing trajectories retains monotonicity. Moreover, We provide a strategy to construct a continuous warping matrix $\boldsymbol{W}_i$ since the elements $\left\{p_{it}\right\}_{t=1}^{n_i}$ are decimals:
\begin{equation}
	\label{Warping_matrix}
	p_{it}=j \Rightarrow \boldsymbol{W}_i^{\left(t,\lfloor j \rfloor  \right)} = \lceil j \rceil -j, \boldsymbol{W}_i^{\left(t,\lceil j \rceil \right)} = j - \lfloor j \rfloor,
\end{equation}
where $p_{it}$ denotes the position of aligned frame in atoms corresponding to frame $t$ in $\boldsymbol{\rm x}_i$, and the symbols $\lfloor \cdot \rfloor$ and $\lceil \cdot \rceil$ represent the floor function and the ceiling function, respectively.
Therefore, the warping matrix $\boldsymbol{W}_i$ can be presented as a function of $\boldsymbol{\beta}_i$
\begin{equation}
	\boldsymbol{W}_i = \boldsymbol{W}_i \left(\boldsymbol{p}_i \left(\boldsymbol{\beta}_i\right) \right) = \boldsymbol{W}_i \left(\boldsymbol{\beta}_i\right).
\end{equation}

\begin{figure}[tb]
	\centering
	\includegraphics[width=\columnwidth]{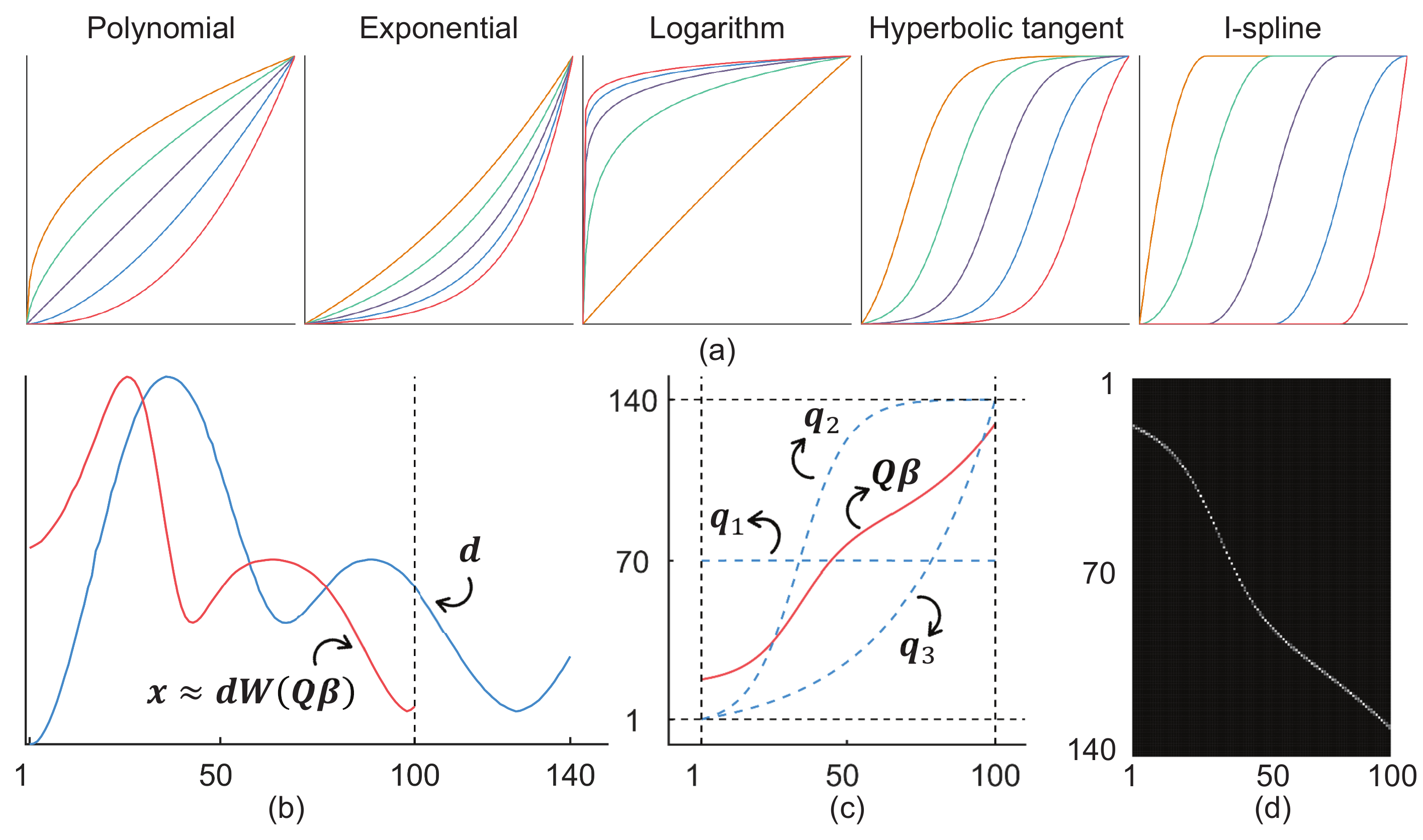}
	\caption{Representing warping path as a combination of monotonic functions. (a) Five
		common choices for monotonic functions. (b) An example of time warping for $\boldsymbol{\rm x}\in \mathbb{R}^{1\times100}$ which is a sub-part of atom $\boldsymbol{d}\in\mathbb{R}^{1\times140}$. (c) The warping function $\boldsymbol{p}=\boldsymbol{Q\beta}$ is a linear combination of three basis functions including a constant function (q1) and two monotonic functions (q2) and (q3). (d) The warping matrix $\boldsymbol{W}$ estimates $\boldsymbol{\rm x}$ as $\boldsymbol{dW}$.}
	\label{GCTW}
\end{figure}

Fig. \ref{GCTW}(a) illustrates five common choices for the basis monotonic functions, including (1) polynomial $\left(ax^b\right)$, (2) exponential $\left(\exp\left(ax+b\right)\right)$, (3) logarithm $\left(\log\left(ax+b\right)\right)$, (4) hyperbolic tangent $\left(\tanh\left(ax+b\right)\right)$ and (5) I-spline \cite{ramsay2005fitting}. A constant basis can be added to the basis set to break the constraints that require warping paths to originate from the bottom-left corner and terminate at the upper-right corner.

Guided by the three types of DTW constraints on the warping path, we impose the following constraints on the coefficients $\boldsymbol{\beta}_i's$. 
\begin{itemize}
	\item Boundary: We enforce the position of the first frame, $p_{i1}=\boldsymbol{q}_i^{\left(1\right)}\boldsymbol{\beta}_i \in \left[1,n_D \gamma\right]$, and the last frame, $p_{in_i}=\boldsymbol{q}_i^{\left(n_i\right)}\boldsymbol{\beta}_i \in \left[n_D \left(1-\gamma\right),n_D\right]$, where $\boldsymbol{q}_i^{\left(1\right)} \in \mathbb{R}^{1\times \tilde{k}}$ and $\boldsymbol{q}_i^{\left(n_i\right)} \in \mathbb{R}^{1\times \tilde{k}}$ are the first and last rows of the basis matrix $\boldsymbol{Q_i}$ respectively and $\gamma$ is a constraint parameter to avoid overcompression in the alignment. 
	\item Monotonicity: We enforce $t_1 < t_2 \Rightarrow p_{it_1}\le p_{it_2}$ by constraining the sign of the coefficients $\boldsymbol{\beta}_i \ge 0 $.
	\item Continuity: We do not impose constraints in this aspect since the $k$ pre-defined continuous basis functions ensure continuity.
\end{itemize}
In contrast to DTW, which imposes a tight boundary (i.e., $p_{i1}=1, p_{in_i}=n_D$), our method allows sub-sequence alignments where the samples can be aligned with a sub-part of atoms. This relaxation is useful when the collected time series reflect part of a complete pattern. For instance, Fig. \ref{GCTW} illustrates an example of matching a shorter sample (red) to a sub-sequence of the longer atom (blue). In this sub-sequence alignment problem, our method models the warping path $\boldsymbol{p}$ as a linear combination of three basis functions including a constant function (q1) and two monotonic functions (q2) and (q3). The warping matrix $\boldsymbol{W}$ is then calculated by Equation (\ref{Warping_matrix}), which satisfies $\boldsymbol{1}_{140}'  \boldsymbol{W} = \boldsymbol{1}_{100}'$.

In summary, we constrain the warping path by adding the following constraints on the coefficients $\boldsymbol{\beta}_i's$,
\begin{equation}
	\boldsymbol{L}_i\boldsymbol{\beta}_i \le \boldsymbol{b}_i,
\end{equation}
where $\boldsymbol{L}_i=\left[-\boldsymbol{I}_{\tilde{k}};-\boldsymbol{q}_i^{\left(1\right)};\boldsymbol{q}_i^{\left(1\right)};\boldsymbol{q}_i^{\left(n_i\right)};-\boldsymbol{q}_i^{\left(n_i\right)}\right]\in\mathbb{R}^{\left(\tilde{k}+4\right) \times \tilde{k}}$ and $\boldsymbol{b}_i = [\boldsymbol{0}_{\tilde{k}}; -1; n_D\gamma;$ $ n_D; -n_D\left(1-\gamma\right)]\in\mathbb{R}^{\tilde{k}}.$ The optimization problem (\ref{opt0}) turns to the following problem
\begin{footnotesize}
\begin{equation}
	\label{opt1}
	\begin{gathered}
		\mathop{\min}\limits_{\left\{\boldsymbol{D}_j\right\}_{j=1}^{p},\left\{\boldsymbol{\alpha}_i\right\}_{i=1}^{m}, \left\{\boldsymbol{\beta}_i \right\}_{i=1}^{m}}  \sum_{i=1}^{m}  \left\{\frac{1}{n_i}\sum_{j=1}^{p}\left \| \boldsymbol{\rm x}_{ij} - \boldsymbol{\alpha}_{i}^{T}\boldsymbol{D}_{j}\boldsymbol{W}_i\left(\boldsymbol{\beta}_i\right)  \right \|_2^2 + \lambda \boldsymbol{1}' \boldsymbol{\alpha}_i \right\} ,\\
		\text{s.t.} \ \ \boldsymbol{\alpha}_{i}\ge\boldsymbol{0}, \boldsymbol{L}_i\boldsymbol{\beta}_i \le \boldsymbol{b}_i, \left\| \boldsymbol{d}_{jk} \right\|_2=1,
	\end{gathered}
\end{equation}
\end{footnotesize}
We name this formulation along with the sequential optimization algorithm as generalized time warping invariant dictionary learning (GTWIDL).

\subsection{Optimization}
The optimization problem (\ref{opt1}) is non-convex with respect to the dictionary atoms $\left\{\boldsymbol{d}_k\right\}_{k=1}^{K}$, the sparseness coefficients $\left\{\boldsymbol{\alpha}_i\right\}_{i=1}^{m}$, and the combination coefficients of warping paths $\left\{\boldsymbol{\beta}_i \right\}_{i=1}^{m}$. In this subsection, we propose an efficient algorithm to solve this problem via the block coordinate descent (BCD) approach. The proposed algorithm consists of two steps, which alternately optimizes the dictionary atoms with two coefficients fixed and optimizes the two coefficients with dictionary atoms fixed.

In the first step, with dictionary atoms $\left\{\boldsymbol{d}_k\right\}_{k=1}^{K}$ fixed, the optimization problem (\ref{opt1}) can be decomposed into $m$ independent sub-problems
\begin{equation}
	\label{opt2}
	\begin{gathered}
		\mathop{\min}\limits_{\boldsymbol{\alpha}_i, \boldsymbol{\beta}_i }  \frac{1}{n_i}\sum_{j=1}^{p}\left \| \boldsymbol{\rm x}_{ij} - \boldsymbol{\alpha}_{i}^{T}\boldsymbol{D}_{j} \boldsymbol{W}_i\left(\boldsymbol{Q}_i \boldsymbol{\beta}_i\right)  \right \|_2^2 + \lambda \boldsymbol{1}' \boldsymbol{\alpha}_i  ,\\
		\text{s.t.} \ \ \boldsymbol{\alpha}_{i}\ge\boldsymbol{0}, \boldsymbol{L}_i\boldsymbol{\beta}_i \le \boldsymbol{b}_i.
	\end{gathered}
\end{equation}
It is worth noting that parallel computing can be easily implemented in these independent sub-problems. To shorten the notation, we denote the term $\boldsymbol{\alpha}_{i}^{T}\boldsymbol{D}_{j} \boldsymbol{W}_i\left(\boldsymbol{Q}_i \boldsymbol{\beta}_i\right)$ as $\boldsymbol{Z}\left(\boldsymbol{\alpha}_i,\boldsymbol{\beta}_i\right) =\left[z_1\left(\boldsymbol{\alpha}_i,\boldsymbol{\beta}_i\right),\ldots,z_{n_i}\left(\boldsymbol{\alpha}_i,\boldsymbol{\beta}_i\right)\right] \in \mathbb{R}^{1\times n_i}$. For the $t$-th frame, we have
\begin{equation}
	z_t\left(\boldsymbol{\alpha}_i,\boldsymbol{\beta}_i\right) = \left[ \boldsymbol{\alpha}_{i}^{T}\boldsymbol{D}_{j} \boldsymbol{W}_i\left(\boldsymbol{Q}_i \boldsymbol{\beta}_i\right) \right]_t = \boldsymbol{\alpha}_{i}^{T}\boldsymbol{D}_{j} \left[ \boldsymbol{W}_i\left(\boldsymbol{Q}_i \boldsymbol{\beta}_i\right) \right]_t , 
\end{equation}
where $[\cdot]_t$ denotes the $t$-th column of a matrix. First consider a simple case where $\boldsymbol{q}_i^{\left(t\right)}\boldsymbol{\beta}_i$ is an integer, where $\boldsymbol{q}_i^{\left(t\right)} \in \mathbb{R}^{1\times k}$ are the $t$-th row of the basis matrix $\boldsymbol{Q_i}$. $\left[ \boldsymbol{W}_i\left(\boldsymbol{Q}_i \boldsymbol{\beta}_i\right) \right]_t$ is a zero vector with only the $\boldsymbol{q}_i^{\left(t\right)}\boldsymbol{\beta}_i$-th element equal 1. Then, $\boldsymbol{\alpha}_{i}^{T}\boldsymbol{D}_{j} \left[ \boldsymbol{W}_i\left(\boldsymbol{Q}_i \boldsymbol{\beta}_i\right) \right]_t$ equals the $\boldsymbol{q}_i^{\left(t\right)}\boldsymbol{\beta}_i$-th frame of $\boldsymbol{\alpha}_{i}^{T}\boldsymbol{D}_{j} $, i.e., $z_t\left(\boldsymbol{\alpha}_i,\boldsymbol{\beta}_i\right) = \left[ \boldsymbol{\alpha}_{i}^{T}\boldsymbol{D}_{j} \right]_{\boldsymbol{q}_i^{\left(t\right)}\boldsymbol{\beta}_i}$. The first-order Taylor expansion of $z_t$ around $\boldsymbol{\beta}_i$ can be represented as
\begin{footnotesize}
\begin{equation}
	\label{Taylor}
	\begin{aligned}
		z_t\left(\boldsymbol{\alpha}_i,\boldsymbol{\beta}_i + \Delta\boldsymbol{\beta}_i \right) &\approx 	z_t\left(\boldsymbol{\alpha}_i,\boldsymbol{\beta}_i \right) + \nabla \left. \left(\boldsymbol{\alpha}_{i}^{T}\boldsymbol{D}_{j}\right)\right|_{\boldsymbol{q}_i^{\left(t\right)}\boldsymbol{\beta}_i} \frac{\partial \boldsymbol{q}_i^{\left(t\right)}\boldsymbol{\beta}_i}{\partial \boldsymbol{\beta}_i} \Delta\boldsymbol{\beta}_i, \\
		& =z_t\left(\boldsymbol{\alpha}_i,\boldsymbol{\beta}_i \right) + \nabla \left. \left(\boldsymbol{\alpha}_{i}^{T}\boldsymbol{D}_{j}\right)\right|_{\boldsymbol{q}_i^{\left(t\right)}\boldsymbol{\beta}_i} \boldsymbol{q}_i^{\left(t\right)} \Delta\boldsymbol{\beta}_i,
	\end{aligned}
\end{equation}
\end{footnotesize}
where $\Delta\boldsymbol{\beta}_i$ denotes a small displacement around $\boldsymbol{\beta}_i$ and $\nabla\left. \left(\boldsymbol{\alpha}_{i}^{T}\boldsymbol{D}_{j}\right)\right|_{\boldsymbol{q}_i^{\left(t\right)}\boldsymbol{\beta}_i}\in\mathbb{R}$ denotes the gradient of the combination of atoms $\boldsymbol{\alpha}_{i}^{T}\boldsymbol{D}_{j}$ around the $\boldsymbol{q}_i^{\left(t\right)}\boldsymbol{\beta}_i$-th frame. The term,
$\frac{\partial \boldsymbol{q}_i^{\left(t\right)}\boldsymbol{\beta}_i}{\partial \boldsymbol{\beta}_i} = \boldsymbol{q}_i^{\left(t\right)} \in \mathbb{R}^{1\times \tilde{k}}$ is the Jacobian of the warping path. Equation (\ref{Taylor}) can be extended to the generalized cases where $\boldsymbol{q}_i^{\left(t\right)}\boldsymbol{\beta}_i$ is a decimal. The gradient term, $\nabla\left. \left(\boldsymbol{\alpha}_{i}^{T}\boldsymbol{D}_{j}\right)\right|_{\boldsymbol{q}_i^{\left(t\right)}\boldsymbol{\beta}_i}$ can be calculated by linear interpolation. Put together the approximations of $z_t\left(\boldsymbol{\alpha}_i,\boldsymbol{\beta}_i + \Delta\boldsymbol{\beta}_i \right)$ and we have
\begin{equation}
	\begin{aligned}
		\boldsymbol{Z}^{T}\left(\boldsymbol{\alpha}_i,\boldsymbol{\beta}_i + \Delta\boldsymbol{\beta}_i \right) \approx \boldsymbol{Z}^{T}\left(\boldsymbol{\alpha}_i,\boldsymbol{\beta}_i  \right) +  \boldsymbol{G}_{ij}^{\beta}\Delta\boldsymbol{\beta}_i,\\
		\text{where } \boldsymbol{G}_{ij}^{\beta}=\left[
		\begin{gathered}
			\nabla \left. \left(\boldsymbol{\alpha}_{i}^{T}\boldsymbol{D}_{j}\right)\right|_{\boldsymbol{q}_i^{\left(1\right)}\boldsymbol{\beta}_i} \boldsymbol{q}_i^{\left(1\right)}\\
			\vdots\\
			\nabla \left. \left(\boldsymbol{\alpha}_{i}^{T}\boldsymbol{D}_{j}\right)\right|_{\boldsymbol{q}_i^{\left(n_i\right)}\boldsymbol{\beta}_i} \boldsymbol{q}_i^{\left(n_i\right)}
		\end{gathered}
		\right]\in\mathbb{R}^{n_i\times \tilde{k}}.
	\end{aligned}
\end{equation}

The first-order Taylor expansion of $z_t$ around $\boldsymbol{\alpha}_i$ can be represented as
\begin{equation}
	\label{Taylor1}
	\boldsymbol{Z}^{T}\left(\boldsymbol{\alpha}_i + \Delta\boldsymbol{\alpha}_i,\boldsymbol{\beta}_i \right) \approx \boldsymbol{Z}^{T}\left(\boldsymbol{\alpha}_i,\boldsymbol{\beta}_i \right) + \boldsymbol{G}_{ij}^{\alpha} \Delta\boldsymbol{\alpha}_{i},
\end{equation}
where $\boldsymbol{G}_{ij}^{\alpha}=\boldsymbol{W}^{T}_i\left(\boldsymbol{Q}_i \boldsymbol{\beta}_i \right) \boldsymbol{D}^{T}_{j}\in\mathbb{R}^{n_i\times K}$. Ignoring the second-order interaction term of $\Delta\boldsymbol{\alpha}_i$ and $\Delta\boldsymbol{\beta}_i$, we have the following approximation
\begin{equation}
	\label{Taylor2}
	\boldsymbol{Z}^{T}\left(\boldsymbol{\alpha}_i + \Delta\boldsymbol{\alpha}_i,\boldsymbol{\beta}_i + \Delta\boldsymbol{\beta}_i\right) \approx \boldsymbol{Z}^{T}\left(\boldsymbol{\alpha}_i,\boldsymbol{\beta}_i \right) + \boldsymbol{G}_{ij}^{\alpha} \Delta\boldsymbol{\alpha}_{i}+  \boldsymbol{G}_{ij}^{\beta}\Delta\boldsymbol{\beta}_i.
\end{equation}

Now we use the Gauss-Newton method to iteratively update $\hat{\boldsymbol{\alpha}}_i = \boldsymbol{\alpha}_i + \Delta\boldsymbol{\alpha}_i$ and $\hat{\boldsymbol{\beta}}_i = \boldsymbol{\beta}_i + \Delta\boldsymbol{\beta}_i$ by plugging Equation (\ref{Taylor2}) into optimization problem (\ref{opt2})
\begin{equation}
	\label{opt3}
	\begin{gathered}
		\mathop{\min}\limits_{ \Delta\boldsymbol{\alpha}_i,  \Delta\boldsymbol{\beta}_i }  \frac{1}{n_i}\sum_{j=1}^{p}\left \|  \boldsymbol{v}_{ij}- \boldsymbol{G}_{ij}^{\alpha} \Delta\boldsymbol{\alpha}_{i}-  \boldsymbol{G}_{ij}^{\beta}\Delta\boldsymbol{\beta}_i  \right \|_2^2 + \lambda \left(\boldsymbol{\alpha}_i+\Delta\boldsymbol{\alpha}_i\right)  ,\\
		\text{s.t.} \ \ \boldsymbol{\alpha}_i+\Delta\boldsymbol{\alpha}_i\ge0, \boldsymbol{L}_i\left(\boldsymbol{\beta}_i+\Delta\boldsymbol{\beta}_i\right) \le \boldsymbol{b}_i,
	\end{gathered}
\end{equation}
where $\boldsymbol{v}_{ij}=\boldsymbol{\rm x}^T_{ij} - \boldsymbol{\alpha}_{i}^{T}\boldsymbol{D}_{j} \boldsymbol{W}_i\left(\boldsymbol{Q}_i \boldsymbol{\beta}_i\right) $. Such an optimization problem can be transformed as a quadratic programming problem
\begin{equation}
	\label{opt4}
	\begin{gathered}
		\mathop{\min}\limits_{ \boldsymbol{\delta}_i }  \frac{1}{2}\boldsymbol{\delta}_i ^T\boldsymbol{H}_i\boldsymbol{\delta}_i  +\boldsymbol{f}_i^T\boldsymbol{\delta}_i ,
		\ \ \text{s.t.} \ \tilde{\boldsymbol{L}}_i\boldsymbol{\delta}_i \le \tilde{\boldsymbol{b}}_i - \tilde{\boldsymbol{L}}_i\boldsymbol{\eta}_i,
	\end{gathered}
\end{equation}
where $\boldsymbol{\delta}_i =\left[\begin{aligned} \Delta\boldsymbol{\alpha}_i\\ \Delta\boldsymbol{\beta}_i	\end{aligned}\right],\  \boldsymbol{\eta}_i=\left[\begin{aligned} \boldsymbol{\alpha}_i\\ \boldsymbol{\beta}_i	\end{aligned}\right],\  \boldsymbol{H}_i=\frac{1}{n_i}\sum_{j=1}^{p}\boldsymbol{G}_{ij}^T\boldsymbol{G}_{ij},\ \boldsymbol{f}_i=-\frac{1}{n_i}\sum_{j=1}^{p}\boldsymbol{G}_{ij}^T\boldsymbol{v}_{ij}+\frac{1}{2}\lambda\left[\boldsymbol{1}_K;\boldsymbol{0}_{\tilde{k}}\right],\ \tilde{\boldsymbol{L}}_i=
\left[\begin{aligned}
	-\boldsymbol{I}_{K},\boldsymbol{0}\\
	\boldsymbol{0},\boldsymbol{L}_i
\end{aligned}\right],\ \tilde{\boldsymbol{b}}_i=\left[\boldsymbol{0}_K;\boldsymbol{b}_i\right],\ \boldsymbol{G}_{ij}=\left[\boldsymbol{G}_{ij}^{\alpha},\boldsymbol{G}_{ij}^{\beta}\right]$.

Such quadratic programming problems can be easily solved and the optimization process turns to solve multiple quadratic programming problems, which is equivalent to the sequential quadratic programming (SQP) method. In this method, the initialization of $\boldsymbol{\alpha}_i$ is randomly sampled from the interval $\left[0,1\right]$ and satisfies the condition $\sum\boldsymbol{\alpha}_i=1$. The initialization of $\boldsymbol{\beta}_i$ is selected to ensure that the generated warping path $\boldsymbol{q}_i$ is a straight line from bottom left to top right without any time warping. The iterative updates of $\boldsymbol{\alpha}_i,\boldsymbol{\beta}_i$ are performed until convergence. The stop conditions are set as
\begin{equation}
	\label{stop1}
	\Delta\boldsymbol{\alpha}_i\le\epsilon_{\alpha},\Delta\boldsymbol{\beta}_i\le\epsilon_{\beta},
\end{equation}
where $\epsilon_{\alpha}$ and $\epsilon_{\beta}$ are pre-specified small positive numbers (e.g., $10^{-3}$).

In the second step, with the coefficients $\boldsymbol{\alpha}_i's, \boldsymbol{\beta}_i's$ fixed, the optimization problem of updating dictionary atoms $\boldsymbol{D}_j$ can be represented as
\begin{equation}
	\label{opt5}
	\begin{gathered}
	\mathop{\min}\limits_{\left\{\boldsymbol{d}_{jk}\right\}_{k=1}^{K}}  \frac{1}{n_i}\sum_{i=1}^{m} \left \| \boldsymbol{\rm x}_{ij} - \sum_{k=1}^{K} \boldsymbol{\alpha}_{ik}\boldsymbol{d}_{jk}\boldsymbol{W}_i\left(\boldsymbol{\beta}_i\right)  \right \|_2^2  ,
	\\\text{s.t.} \ \left\| \boldsymbol{d}_{jk} \right\|_2=1.
	\end{gathered}
\end{equation}
We iteratively update the $k$-th atom $\boldsymbol{d}_{jk}$ with the other atoms fixed. The update formula can be easily deduced by derivation
\begin{equation}
	\hat{\boldsymbol{d}}_{jk} = \boldsymbol{d}_{jk} + \left(\sum_{i=1}^{m}\alpha_{ik}\boldsymbol{u}_{ik}\boldsymbol{W}_i^T\right)\left(\sum_{i=1}^{m}\alpha_{ik}^2\boldsymbol{W}_i\boldsymbol{W}_i^T\right)^{-1},
\end{equation}
where $\boldsymbol{u}_{ik}=\boldsymbol{\rm x}_{ij} - \sum_{k'\ne k } \boldsymbol{\alpha}_{ik'}\boldsymbol{d}_{jk'}\boldsymbol{W}_i\left(\boldsymbol{\beta}_i\right)$ denotes the residual without the $k$-th atom. However, the existence of the inverse term is not guaranteed due to the potential occurrence of a warping matrix $\boldsymbol{W}_i$ with zeros at the start and end frames. Therefore, we approximate this update formula with
\begin{equation}
	\label{updateD}
	\hat{\boldsymbol{d}}_{jk} = \boldsymbol{d}_{jk} + \frac{1}{\sum_{i=1}^{m}\alpha_{ik}^2} \left(\sum_{i=1}^{m}\alpha_{ik}\tilde{\boldsymbol{u}}_{ik}\right) ,
\end{equation}
where $\tilde{\boldsymbol{u}}_{ik}=\mathcal{F}\left(\boldsymbol{\rm x}_{ij}\tilde{\boldsymbol{W}}_i\left(\boldsymbol{\beta}_i\right)\right) - \sum_{k'\ne k } \boldsymbol{\alpha}_{ik'}\boldsymbol{d}_{jk'}$ and the missing value imputation function $\mathcal{F}(\cdot)$ is used to fill the leading and trailing missing values of $\boldsymbol{\rm x}_{ij}\tilde{\boldsymbol{W}}_i\left(\boldsymbol{\beta}_i\right)$ with their nearest non-missing value. Here, $\tilde{\boldsymbol{W}}_i$ denotes the inverse warping matrix which warps time series $\boldsymbol{\rm x}_{ij}$ to the space of dictionary. 
%This update formula is the solution to the transformed optimization problem
%\begin{equation}
%	\label{opt6}
%	\begin{gathered}
	%		\mathop{\min}\limits_{\left\{\boldsymbol{d}_{jk}\right\}_{k=1}^{K}}  \frac{1}{n_i}\sum_{i=1}^{m} \left \| \text{fillmissing}\left(\boldsymbol{\rm X}_{ij}\tilde{\boldsymbol{W}}_i\left(\boldsymbol{\beta}_i\right)\right) - \sum_{k=1}^{K} \boldsymbol{\alpha}_{ik}\boldsymbol{d}_{jk}  \right \|_2^2  ,\\
	%		\text{s.t.} \ \left\| \boldsymbol{d}_{jk} \right\|_2=1.
	%	\end{gathered}
%\end{equation}

In this method, the initialization of $\boldsymbol{d}_{jk}'s$ is set as the first $\tilde{k}$ largest eigenvectors based on the samples after stretching. The stop conditions are set as
\begin{equation}
	\label{stop2}
	\left\|\hat{\boldsymbol{d}}_{jk}-\boldsymbol{d}_{jk}\right\|_2^2 \le\epsilon_{\boldsymbol{d}},\ k=1,\ldots,K,
\end{equation}
where $\epsilon_{\boldsymbol{d}}$ is pre-specified small positive numbers (e.g., $10^{-2}$). Similar conditions are set for the relative difference of successive estimates as stopping conditions for the block coordinate descent processes. The optimization algorithm is summarized in Algorithm 1.

\begin{algorithm}
	\caption{GTWIDL Algorithm}
	\SetKwInOut{KwIn}{Input}
	\SetKwInOut{KwOut}{Output}
	\KwIn{$\left\{\boldsymbol{X}_i\right\}_{i=1}^{m},\lambda,K$}
	\KwOut{$\left\{\boldsymbol{D}_j\right\}_{j=1}^{p},\left\{\boldsymbol{\alpha}_i\right\}_{i=1}^{m}, \left\{\boldsymbol{\beta}_i \right\}_{i=1}^{m}$}
	\textbf{Initialization:} Initial $\left\{\boldsymbol{D}_j\right\}_{j=1}^{p},\left\{\boldsymbol{\alpha}_i\right\}_{i=1}^{m}, \left\{\boldsymbol{\beta}_i \right\}_{i=1}^{m}$ and $\left\{\boldsymbol{Q}_i \right\}_{i=1}^{m}$ 
	\Repeat{convergence}{
		\ForPar{$i=1,\ldots,m$}{
			\Repeat{convergence}{
				calculate $\boldsymbol{H}_i,\boldsymbol{f}_i,\tilde{\boldsymbol{L}}_i,\tilde{\boldsymbol{b}}_i$\;
				update $\hat{\boldsymbol{\alpha}}_i \leftarrow \boldsymbol{\alpha}_i + \Delta\boldsymbol{\alpha}_i$ and $\hat{\boldsymbol{\beta}}_i \leftarrow \boldsymbol{\beta}_i + \Delta\boldsymbol{\beta}_i$ by solving (\ref{opt4})\;
			}
		}
		\ForPar{$j=1,\ldots,p$}{
			\Repeat{convergence}{
				\For{k=1,$\ldots$,K}{
					calculate $\tilde{\boldsymbol{u}}_{ik}$\;
					update $\hat{\boldsymbol{d}}_{jk} \leftarrow \boldsymbol{d}_{jk} + \frac{1}{\sum_{i=1}^{m}\alpha_{ik}^2} \left(\sum_{i=1}^{m}\alpha_{ik}\tilde{\boldsymbol{u}}_{ik}\right) $\;}
			}
		}
	}
\end{algorithm}

\subsection{Complexity analysis}
\label{sec33}

The time complexity of the GTWIDL algorithm consists of two parts: coefficients update and dictionary update. In the coefficient update part, the time cost of one loop is $O\left(\left(\tilde{k}+K\right)^2 m n_d + \left(\tilde{k}+K\right)^3 m \right)$. The first term relates to the calculation of $\boldsymbol{H}_i,\boldsymbol{f}_i,\tilde{\boldsymbol{L}}_i,\tilde{\boldsymbol{b}}_i$ on $m$ samples, and the second term is associated with solving the quadratic problem. In the dictionary update part, the time cost of one loop is $O\left(\tilde{k} m n_d + K m n_d\right)$. Here, the first term accounts for the calculation of $\tilde{\boldsymbol{W}}_i$ on $m$ samples, and the second term relates to the dictionary update based on Equation (\ref{updateD}). Since $\left(\tilde{k}+K\right) \ll n_d$, the computational complexity for one iteration is $O\left(\left(\tilde{k}+K\right)^2 m n_d\right)$. Denoting $T_1,T_2$ as the maximum iteration of BCD update and coefficients update separately, the total time complexity of the GTWIDL algorithm is $O\left(T_1 T_2\left(\tilde{k}+K\right)^2 m n_d\right)$. It is demonstrated in our case studies that the
GTWIDL algorithm converges fast, where $T_1$ can be set from 5 to 50 and $T_2$ can be set from 5 to 20.

\begin{table}[tbp]
	\caption{Comparison of dictionary learning algorithms in degrees of freedom and complexity.}
	\label{Tab1}
	\centering
	%\resizebox{\columnwidth}{!}{
		\begin{tabular}{r|lll|l}
			\hline
			\multirow{2}{*}{Method} & \multicolumn{3}{c|}{Degrees of Freedom}                                                  & \multicolumn{1}{c}{\multirow{2}{*}{Complexity $O(\cdot)$}} \\ \cline{2-4}
			& \multicolumn{1}{c|}{Warping} & \multicolumn{1}{c|}{Coding} & \multicolumn{1}{c|}{Learning} & \multicolumn{1}{c}{}                            \\ \hline
			TWI-$k$SVD                & \multicolumn{1}{l|}{$mpn_D$}        & \multicolumn{1}{l|}{$mK$}       &     $pKn_D$                          &       $T_1 \left(T_2 m n_d^2 + K n_d^3\right) $                                            \\
			RISLDTW                 & \multicolumn{1}{l|}{$mn_D$}        & \multicolumn{1}{l|}{$mpK$}       &       $pKn_D$            &    $T_1 \left(T_2 m n_d^2 + p n_d^3\right) $            \\
			GTWIDL                  & \multicolumn{1}{l|}{$m\tilde{k}$}        & \multicolumn{1}{l|}{$mK$}       &     $pKn_D$                &         $T_1 T_2\left(\tilde{k}+K\right)^2 m n_d$                    \\ \hline
		\end{tabular}
		%}
\end{table}

\begin{figure}[tbhp]
	\centering
	\includegraphics[width=0.7\columnwidth]{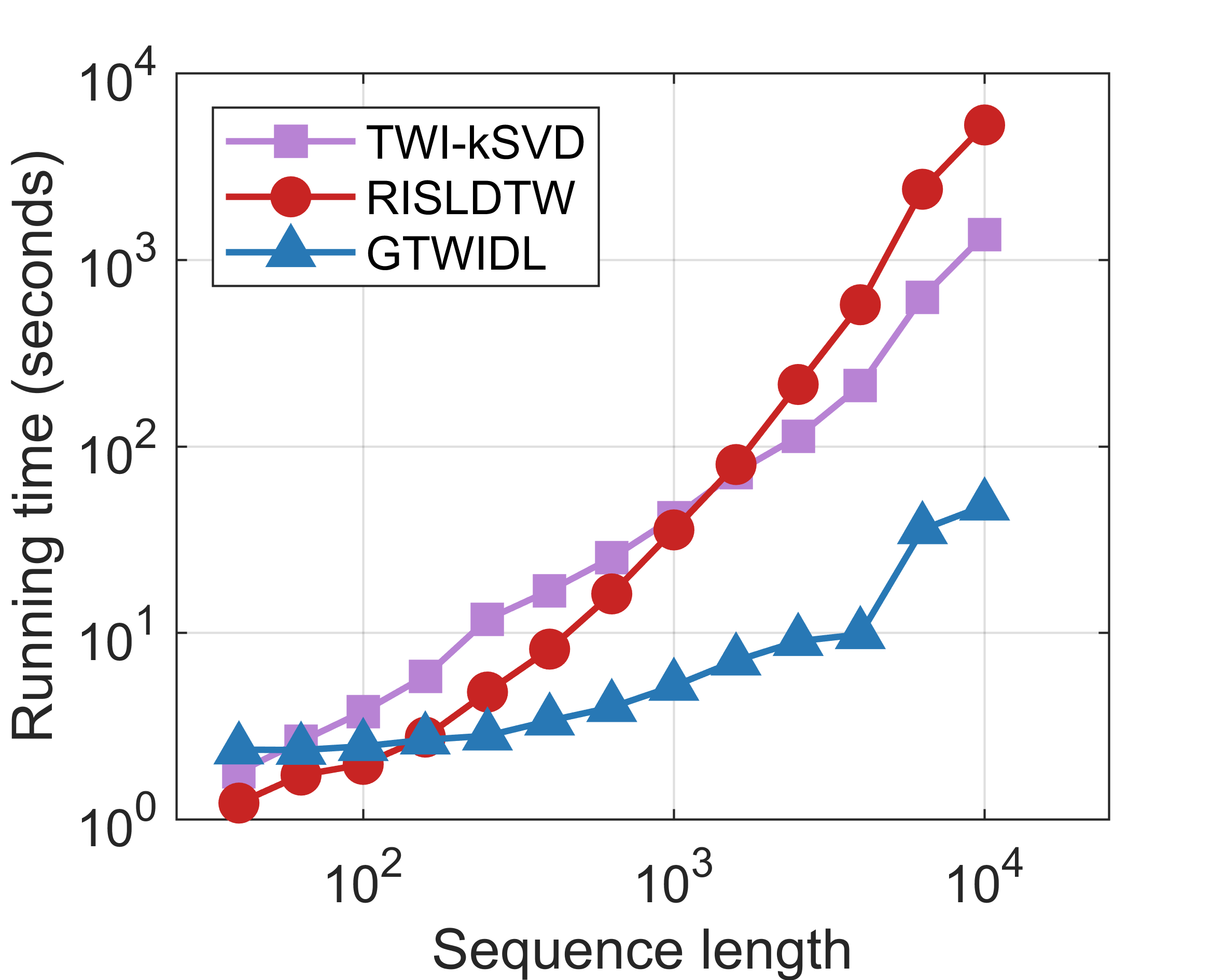}
	\caption{Simulation of the running time (in seconds) for dictionary learning on a dataset containing 10 samples.}
	\label{time_complexity}
\end{figure}

The comparison of degrees of freedom and complexity between existing dictionary learning algorithms \cite{deng2020invariant,yazdi2018time} and the GTWIDL is provided in Tab. \ref{Tab1}. By utilizing a linear combination of basis functions to approximate the warping path, our proposed method requires fewer optimization parameters compared to other approaches. In addition, the GTWIDL method offers superior computational efficiency due to its linear complexity with respect to the sequence length. In contrast, other methods require quadratic time complexity to calculate the DTW warping path and cubic time complexity to implement PCA or SVD for dictionary updating. The running time comparison in a simulation study is illustrated in Fig. \ref{time_complexity}. As the increase of sequence length, TWI-$k$SVD \cite{yazdi2018time} and RISLDTW \cite{deng2020invariant} require much more running time compared with the proposed GTWIDL method.

\subsection{Uniqueness and connections to previous works}

The proposed GTWIDL method distinguishes itself with existing methods in following aspects. First, the GTWIDL incorporates the framework of GCTW \cite{zhou2015generalized} to improve learning performance and applicability by relaxing constraints and boosting optimization procedures. This makes GTWIDL superior to DTW based  dictionary learning methods such as TWI-$k$SVD \cite{yazdi2018time} and RISLDTW \cite{deng2020invariant}. Second, the GTWIDL differs with GCTW by incorporating dictionary learning, which provides more comprehensive investigation of time series data for modeling, classification, and clustering. This is very important contribution of our proposed method since it is the first work that provides a general framework for integrating GCTW with dictionary learning, which demonstrates great potential for leading future research in this area. 

On the other hand, the proposed GTWIDL method also shares some connection with existing methods. For example, when the number of dictionary atoms is set to one, the GTWIDL method resembles global averaging methods under DTW \cite{petitjean2011global}, which aim to obtain an average atom. In this case, the GTWIDL method becomes a centroid problem with a scaling coefficient, and GTWIDL-based classification and clustering methods are similar to centroid-based methods discussed in Section \ref{TWICC}. Therefore, global averaging methods can be regarded as special cases of our algorithm. In contrast, the GTWIDL algorithm can adaptively learn the number of atoms, which is more robust for multiple patters in a class.

\section{Classification and clustering based on GTWIDL}
\label{sec4}
In this section, we propose classification and clustering algorithms tailored for the proposed GTWIDL, where the basic idea is to assign test samples to the class whose dictionary yields the minimum reconstruction error. 

\subsection{Classification based on GTWIDL}
Various approaches have been proposed for using dictionary learning algorithms in classification tasks. One such method \cite{deng2020invariant} involves training a separate dictionary for each class and then combining them into a global dictionary. This global dictionary is then used to sparsely code both training and test samples with subsequent application of classifiers such as 1NN and SVM based on the sparse coefficients. Another approach \cite{yazdi2018time,wright2008robust}, based on the same global dictionary, evaluates test samples by assigning them to the class with the lowest reconstruction error resulting from the dictionary and sparse coefficients. However, the local dictionaries may share similar atoms across different classes. Therefore, the global dictionary may suffer homogeneous atoms and produce non-unique optimal sparse coding results. In such cases, the sparse coefficients and reconstruction error under the global dictionary may not be sufficiently accurate and discriminative for the classification task.

In this subsection, we propose a classification strategy that focuses on the local dictionary per class. First, all dictionaries are learned using training samples. Then, for each test sample, we optimize problem (\ref{opt2}) once per class. This is done to obtain the warping paths and sparse coding of the sample under all dictionaries. Subsequently, we calculate the reconstruction error by $\sum_{j=1}^{p}\left \| \boldsymbol{\rm x}_{ij} - \boldsymbol{\alpha}_{i}^{T}\boldsymbol{D}_{j} \boldsymbol{W}_i\left(\boldsymbol{Q}_i \boldsymbol{\beta}_i\right)  \right \|_2^2 $. Finally, the test sample is assigned to the class exhibiting the minimum reconstruction error.

The complexity of such a classification algorithm is linear with respect to the sequence length, making it superior to other time warping invariant classification methods. A notable advantage of our approach is that it does not need to implement DTW for every pair of training and test samples, as required by DTW distance-based methods. Instead, our algorithm only requires the assignment of each test sample to a few atoms, rendering it more efficient for large datasets.

One potential concern in the proposed classification method is that the reconstruction error is sensitive to the number of atoms $K$ for each dictionary. We resolve this concern by providing an adaptive strategy to select a suitable $K$ for different classes. This approach is superior to setting fixed $K$ for all classes as it considers complicated interactions among internal structures within each class. We first learn a dictionary with enough atoms to fully fit the training samples.  Subsequently, we apply Singular Value Decomposition (SVD) on the warped time series to obtain the eigenvalues, denoted as $\lambda_1\ge\ldots\ge\lambda_m$. The value of $K$ is then chosen such that $\frac{\sum_{k=1}^{K}\lambda_i}{\sum_{k=1}\lambda_i}\ge \zeta$ and $\frac{\sum_{k=1}^{K-1}\lambda_i}{\sum_{k=1}\lambda_i} < \zeta$, where $\zeta$ reflects how well the dictionary fits the data. The parameter $\zeta$, along with $\lambda$, is further selected through cross-validation. The details of this tuning process can be found in Section \ref{tunning}. In our experiment comparisons, this adaptive strategy shows better fit of the dictionary to the internal structure of each class, leading to improved classification accuracy.

\subsection{Clustering based on GTWIDL}
In this subsection, we present a clustering method developed based on the above classification framework. First, we employ a traditional clustering method, such as Sparse Subspace Clustering (SSC, \cite{elhamifar2013sparse}), without temporal warpings, to group the data. In the following step, dictionaries for all the classes are learned by the proposed GTWIDL algorithm. Subsequently, the samples are reassigned to the appropriate class based on the dictionary that yields the minimum reconstruction error. We repeat the second and third steps until convergence. For this task, we set the parameters of the GTWIDL method to their default values, i.e., $\lambda=0.0001$ and $K=5$.

\section{Experiments}
\label{sec5}
In this section, the performance of the proposed GTWIDL method is evaluated from three perspectives: (1) reconstruction to demonstrate its superior representation ability; (2) classification; and (3) clustering to investigate its discriminative ability. All experiments were conducted using Matlab on an Intel Xeon Processor E5-4627 v4 with 128GB RAM, ensuring robust and reliable results.

\subsection{Datasets}
The research experiments were performed on a total of ten datasets, which comprise four public handwritten character datasets, two synthetic datasets, and four time series datasets from the UCR archive \cite{UCRArchive2018}. The four public handwritten character datasets used included {\small DIGITS}, {\small LOWER}, {\small UPPER}, and {\small CHAR-TRAJ}. These datasets involve naturally multivariate time series of varying lengths \cite{chen20126dmg}, and showcase the 2-dimensional handwritten character trajectory of air-handwritten motion gestures of digits, upper and lower case letters, and other characters. Furthermore, two synthetic datasets, namely {\small BME} and {\small UMD}, were employed, which contained time series exhibiting local temporal features within classes while demonstrating distinctive global patterns. In addition, four UCR time series datasets with varying delays were chosen, including {\small ArrowHead}, {\small DSR} (short for {\small DiatomSizeReduction}), {\small FiftyWords}, and {\small Trace}. All these datasets were subject to misalignment to some extent.

\begin{figure}[tbhp]
	\centering
	\includegraphics[width=\columnwidth]{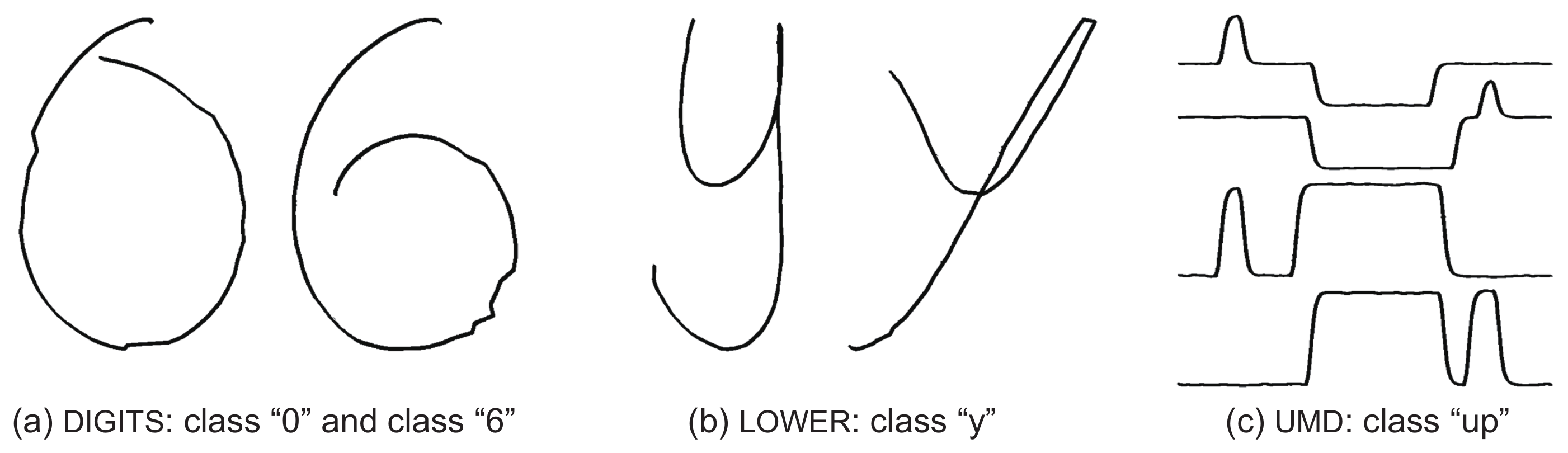}
	\caption{Time series characteristics within and between classes. (a) Examples of class "0" and class "6" in dataset DIGITS. (b) Examples of class "y" in dataset LOWER. (c) Examples of class "up" in dataset UMD.}
	\label{example_datasets}
\end{figure}

Fig. \ref{example_datasets} illustrates the characteristics of the considered time series through several examples. For instance, time series may share similar global patterns between different classes, such as “0” and “6”. Time series within the same class, such as “y”, may exhibit similar global patterns but different shapes representing different writing styles. Time series may even have different global patterns, as in the case of the “up” class, while sharing only local events (such as the “small bell”) that may occur at different time stamps.

In the experiments of dictionary learning and classification, for a fair comparison, the training and testing data are split based on the definition in the dataset when available. Otherwise, we randomly selected ten samples from each class to constitute the training dataset, with the remaining samples serving as the testing dataset. This random selection process was repeated ten times to ensure robustness of our results. In the experiments on clustering, we utilized the aforementioned training datasets as our target datasets. The data characteristics are given in Tab. \ref{Tab2}. In addition, we use a zero-padding strategy \cite{yazdi2018time} to circumvent the problem of variable lengths for some methods demanding a uniform length and splice multidimensional time series into one-dimensional time series for univariate methods.

\begin{table}[tbp]
	\caption{Data description.}
	\label{Tab2}
	\centering
	\resizebox{\columnwidth}{!}{
		\begin{tabular}{rccccc}
			\hline
			Dataset             & Nb. class & Dimensions & Train size & Test size & Length range \\ \hline
			DIGITS              & 10  & 2        & 100        &  500      & 29$\sim$218  \\
			LOWER               & 26  & 2        & 260        &  1210     & 27$\sim$263  \\
			UPPER               & 26  & 2        & 260        &  6241     & 27$\sim$412  \\
			CHAR-TRAJ           & 20  & 3        & 200        &  2658     & 109$\sim$205 \\
			UMD                 & 3   & 1        & 360        &  1440     & 150          \\
			BME                 & 3   & 1        & 300        &  1500     & 128          \\
			ArrowHead                  & 3   & 1        & 36         &  175      & 251          \\
			DSR                 & 4   & 1        & 16         &  306      & 345          \\
			FiftyWords                 & 50  & 1        & 450        &  455      & 270          \\
			Trace                      & 4   & 1        & 100        &  100      & 275          \\
			\hline
		\end{tabular}
		}
\end{table}

\subsection{Benchmarks}
\label{tunning}
In the task of dictionary learning, we use a traditional dictionary learning method (DL) without temporal warping and two time warping invariant dictionary learning methods, i.e., TWI-$k$SVD \cite{yazdi2018time} and RISLDTW \cite{deng2020invariant}, as the benchmarks. 

In the task of time series classification, a variant of RISLDTW namely RISLDTW-ELS, is added which further uses a feature selection model called Efficiently Learning Shapelets (ELS, \cite{hou2016efficient}) to improve accuracy. Additionally, three distance-based methods are also set as benchmarks. They respectively calculate DTW distance \cite{berndt1994using}, DDTW distance \cite{2002Derivative} and WDTW distance \cite{jeong2011weighted} between training and testing samples. A 1NN classifier is then applied to provide final classification results. For each method, parameters are learned via grid search using three-fold cross-validation on the training dataset. Details of parameter candidate sets are provided in Tab. \ref{Tab3}. 

In the task of time series clustering, traditional sparse subspace clustering, the variant of TWI-$k$SVD called TWI-DLCLUST, the variant of RISLDTW, and the three DTW distance-based methods are used as benchmarks. For the RISLDTW method, the k-means clustering method is applied for clustering based on the learned sparse coding. For the DTW distance-based methods, clustering results are achieved via spectral clustering method based on the learned distances, where the similarity matrix is calculated by $S_{ij}=\exp\left(-\frac{d_{ij}}{\sigma^2}\right)$ and $\sigma$ is set to 5 in the following experiments.

\begin{table}[tbp]
	\caption{The candidate sets of parameters in the classification framework.}
	\label{Tab3}
	\resizebox{\columnwidth}{!}{
		\begin{tabular}{llll}
			\hline
			& Para.              & Ranges                               & Description                    \\ \hline
			WDTW        & $g$                        & {[}0.01,0.02,0.03,0.05,0.08,0.1{]}   & Weighted coefficient           \\
			TWI-$k$SVD    & $K$                        & 10                                   & Dictionary size                \\
			& $\tau$     & 2                                    & Sparsity constraint            \\
			RISLDTW     & $c$                        & {[}0,0.1,0.2,0.3,0.4,0.5,0.6{]}         & Variance proportion \\
			RISLDTW-ELS & $\alpha_1$ & {[}0,0.01,0.02,0.04,0.06,0.08,0.1{]} & Regularisation                 \\
			& $\alpha_2$ & {[}0,0.01,0.02,0.03,0.04,0.05{]}     & Regularisation                 \\
			GTWIDL      & $\lambda$   & {[}0.001,0.0005,0.0001,0.00005,0{]}  & Sparsity constraint                  \\
			& $\zeta$     & {[}0.5,0.7,0.8,0.9,0.95,0.99{]}      & Proportion threshold           \\ \hline
	\end{tabular}}
\end{table}

\subsection{Evaluation on dictionary learning}
%\label{sec53}
In this subsection, we first investigate the convergence properties of the proposed GTWIDL algorithm. Here we set two metrics to evaluate the degree of convergence, i.e., reconstruction error and alignment error. Specifically, the reconstruction error is measured by Mean Squared Error (MSE), which was defined as
\begin{equation}
	{\rm MSE} = \frac{1}{n_i}\sum_{j=1}^{p}\left \| \boldsymbol{\rm x}_{ij} - \boldsymbol{\alpha}_{i}^{T}\boldsymbol{D}_{j} \boldsymbol{W}_i\left(\boldsymbol{Q}_i \boldsymbol{\beta}_i\right)  \right \|_2^2.
\end{equation}
The alignment error is measured by the sum-of-pair Euclidean distance between the estimated alignment path and the ground truth, which was defined as
\begin{equation}
	{\rm Error_{align}} = \frac{1}{n_i}\left \| \boldsymbol{W}_i\left(\boldsymbol{Q}_i \boldsymbol{\beta}_i\right) - \boldsymbol{W}_i^t \right \|_2^2,
\end{equation}
where $\boldsymbol{W}_i^t$ denotes the ground truth of the alignment path. When the ground truth is unavailable, we use the final alignment path obtained after 100 iterations as a reference.

\begin{figure}[tbhp]
	\centering
	\includegraphics[width=\columnwidth]{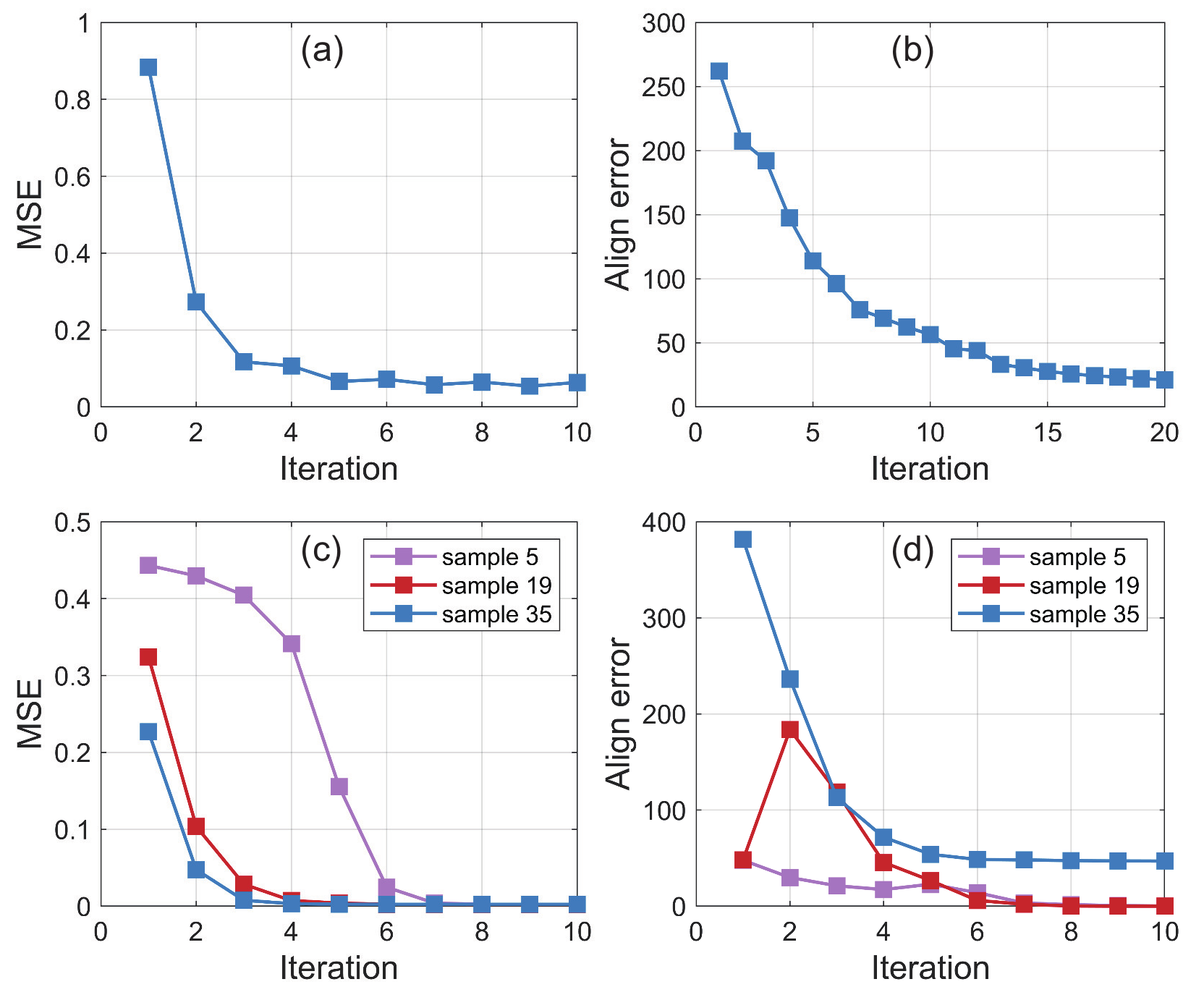}
	\caption{The reconstruction error and the alignment error against the number of iterations on the first class of the Trace dataset. The top row displays the average iteration curves when updating the dictionary, and the bottom row shows the iteration curves when updating coefficients with a fixed dictionary.}
	\label{dl_mse}
\end{figure}

Fig. \ref{dl_mse} illustrates the reconstruction error and the alignment error against the number of iterations on the first class of the Trace dataset. Fig. \ref{dl_mse}(a-b) show the average iteration curves when updating the dictionary, which corresponds to the outer iteration loop in the GTWIDL algorithm. The reconstruction error converges rapidly around the 3rd iteration, which implies that the proposed algorithm could efficiently handle the misaligned data and fit the samples well. The alignment error converges more slowly, around the 15th iteration. This is because the optimal alignment error is often not unique and different paths may have similar alignment performance. For example, as we can see from Fig. \ref{dl_mse}(c-d), for sample 35 of the Trace dataset, our algorithm converges around the 4th iteration with the reconstruction error close to zero. However, the alignment error converges around value 50, which is much larger than zero. It indicates that the alignment path at the 4th iteration differs from the ground truth but still has excellent alignment performance. Fig. \ref{dl_mse}(c-d) shows the iteration curves when updating coefficients with a fixed dictionary, which corresponds to the inner iteration loop in the GTWIDL algorithm. We test sample 5,19 and 35 from the first class of the Trace dataset. Both the reconstruction error and alignment error converge rapidly around the 5th iteration. As a result, we demonstrate that the proposed GTWIDL algorithm has great convergence properties and is capable of efficiently finding the dictionary and aligning samples. We set $T_1$ from 5 to 50 and set $T_2$ from 5 to 20 to achieve such performance, which is discussed in Section \ref{sec33}.

\begin{figure}[tbhp]
	\centering
	\includegraphics[width=\columnwidth]{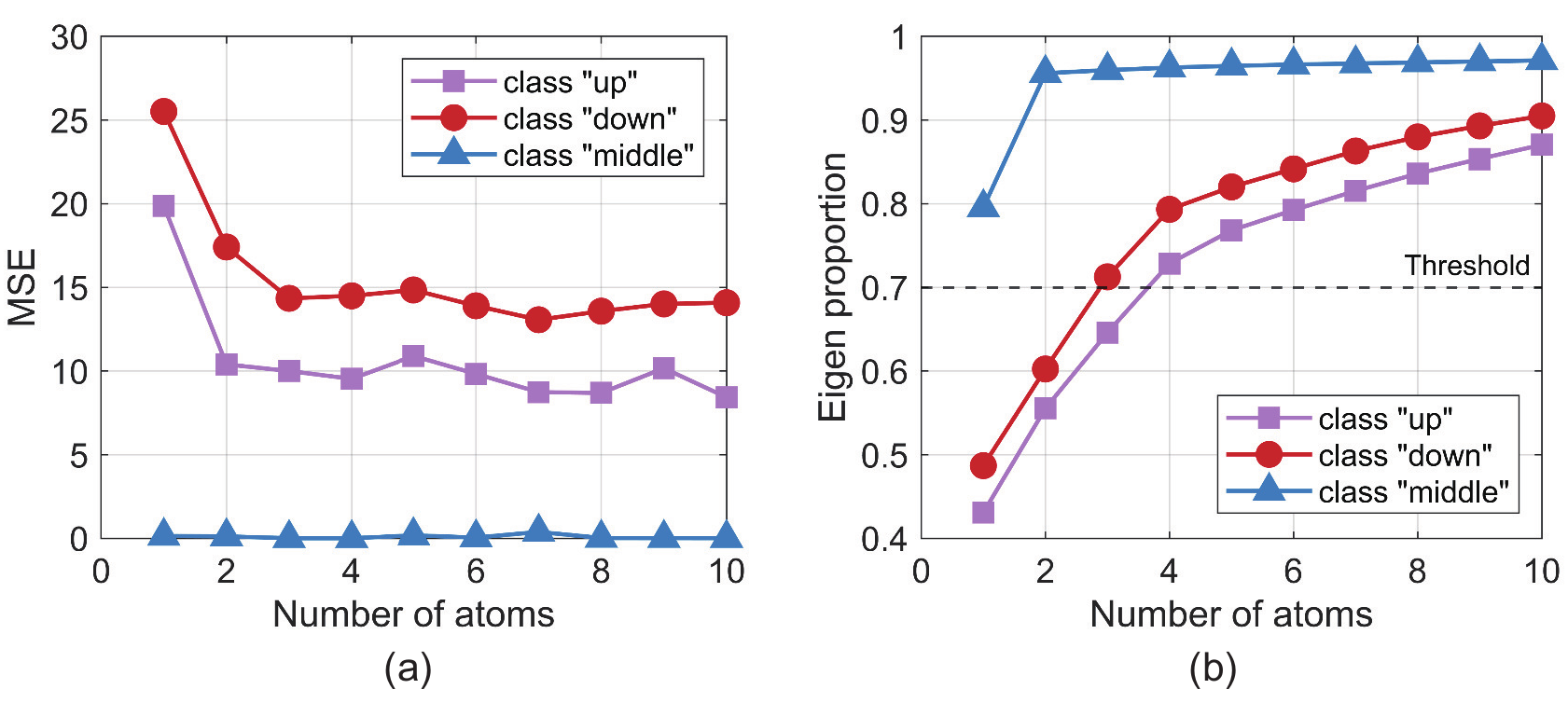}
	\caption{(a) The reconstruction error against the number of atoms for the UMD dataset. (b) The cumulative proportion of eigenvalues based on the samples after alignments.}
	\label{dl_dimension}
\end{figure}

To investigate the influence of the number of atoms on the proposed algorithm, we test the GTWIDL algorithm on the {\small UMD} dataset, which contains several different patterns in each class. For example, the ``up'' and ``down'' classes both have four different patterns, while the ``middle'' class has only one pattern. Fig. \ref{dl_dimension}(a) illustrates the reconstruction error against the number of atoms. The MSE value for the ``up'' and ``down'' classes converge when the number of atoms equals four, while the MSE value for the ``middle'' class converges at the first point. To avoid overfitting and provide refined atoms, the number of atoms should not be too large and should be set around the number of underlying patterns. As introduced in Section 4.1, by setting the threshold $\zeta$ at approximately 0.7 (shown in Fig. \ref{dl_dimension}(b)), the optimal value for the dictionary's dimension can be obtained. However, it is important to note that the optimal thresholds may vary depending on the dataset being analyzed. To accommodate such variations, we provide a candidate list for 3-fold cross-validation to select the threshold in the classification framework, which is displayed in Tab. \ref{Tab3}.

\begin{figure*}[bthp]
	\centering
	\includegraphics[width=0.95\textwidth]{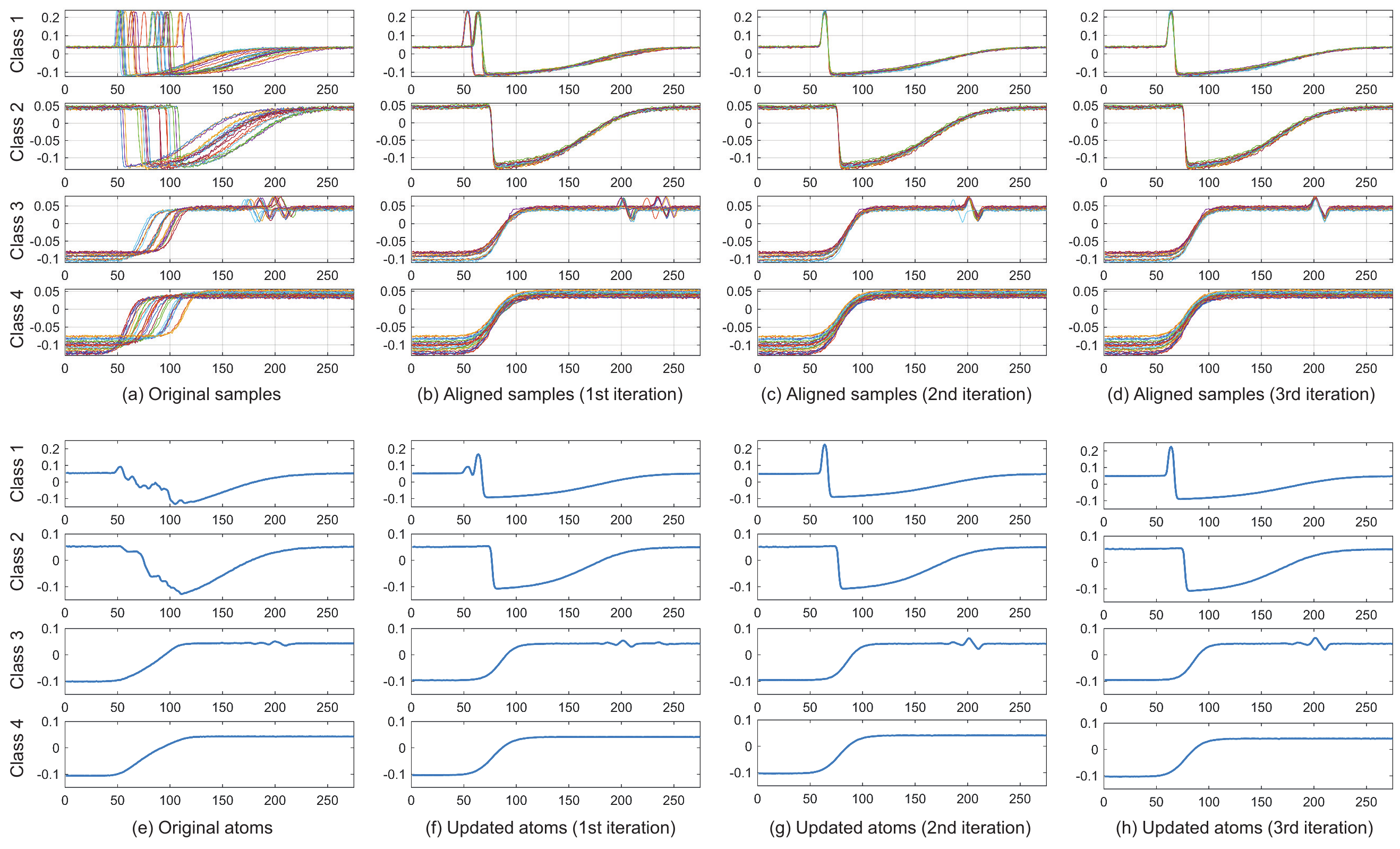}
	\caption{The top and the bottom rows visualize the aligned samples and the updated atoms for each class of the Trace dataset at the initialization, after the 1st iteration, the 2nd iteration and the 3rd iteration, respectively.}
	\label{dl_atoms}
\end{figure*}

Fig. \ref{dl_atoms} illustrates the progress of applying the proposed GTWIDL algorithm to the Trace dataset. The initial misaligned samples belonging to four different patterns are shown in Fig. \ref{dl_atoms}(a). The atoms learned directly from the misaligned data (Fig. \ref{dl_atoms}(e)) fail to capture the local patterns accurately. This is particularly clear for class 1 and class 2, where the learned atoms tend to average the misaligned samples, resulting in abrupt rises and falls. After the first iteration, the data are roughly aligned (Fig. \ref{dl_atoms}(b)), and the simple patterns of class 2 and class 4 can be learned accurately (Fig. \ref{dl_atoms}(f)). With an increasing number of iterations, the alignments of samples become more precise. After the third iteration (Fig. \ref{dl_atoms}(d)), all samples from the four classes are well-aligned. Furthermore, the learned atoms increasingly capture the local patterns and accurately represent the characteristics of different samples. Among the four classes, samples from class 3 are the most challenging to align because they have two key positions for alignment. Specifically, the first key position lies around frame 80 with a rise, and the second lies around frame 200 with a Z-shape wave. This indicates that when samples share more local patterns that require more complex warping paths, it becomes difficult for the GTWIDL algorithm to learn dictionaries and align the samples. This is reasonable since such warping paths have more local features and it takes more iterations to provide accurate estimation, especially when we only use a few basis paths in our GTWIDL algorithm.

\begin{figure*}[bthp]
	\centering
	\includegraphics[width=0.95\textwidth]{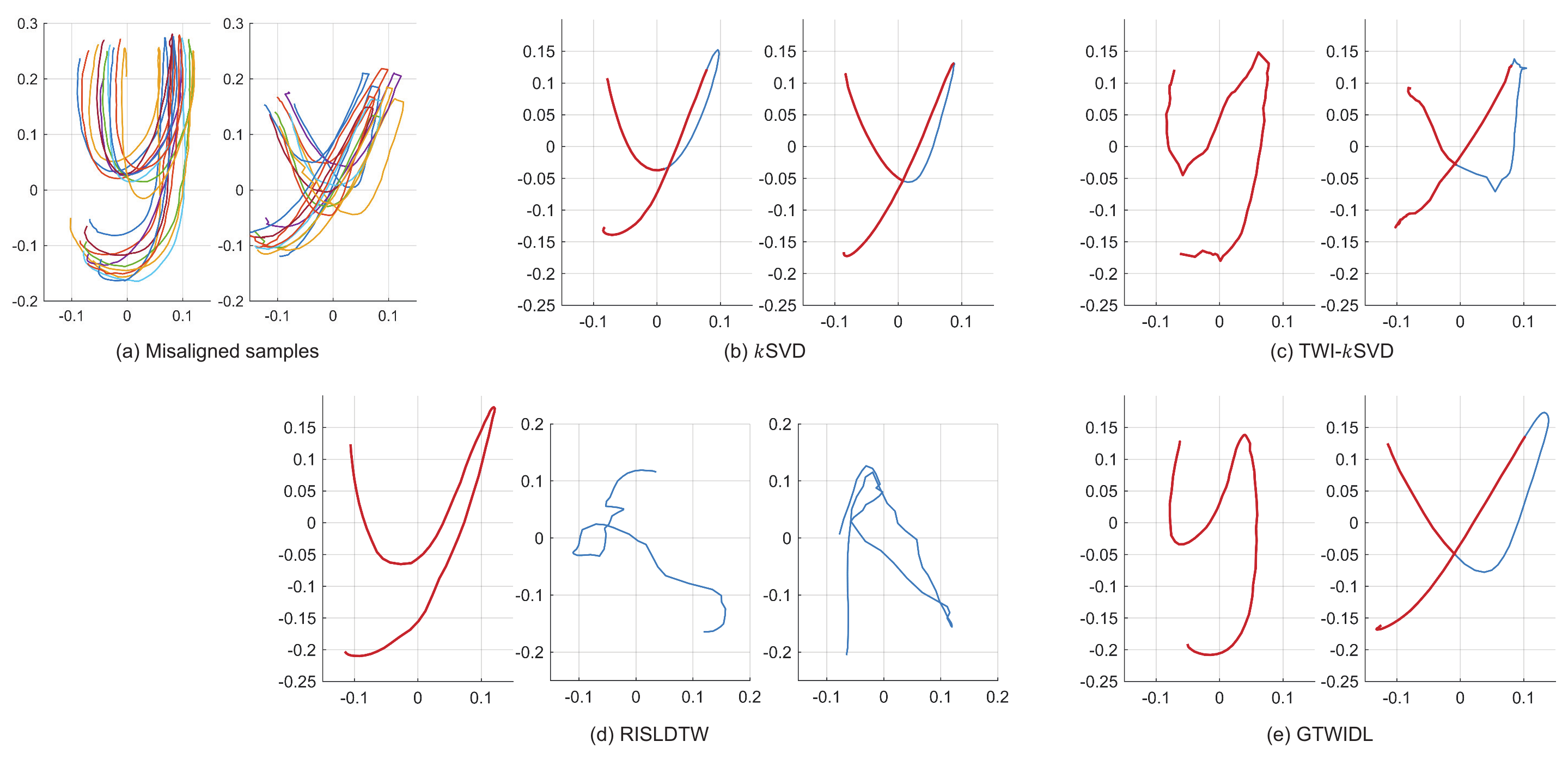}
	\caption{(a) Misaligned samples from `y' class in the LOWER dataset and (b-e) learned atoms via four dictionary learning methods. The red lines of atoms represent the writing rendered on the paper and the blue lines represent the writing rendered in the air or the writing lacking interpretable patterns.}
	\label{dl_dictionary}
\end{figure*}

To evaluate the performance of dictionary learning, we test the GTWIDL algorithm on the two-dimensional {\small LOWER} handwriting dataset and compare the learned atoms with other benchmarks, as shown in Fig. \ref{dl_dictionary}. The misaligned samples from the `y' class have two patterns representing two writing styles. Therefore, we learn two atoms for each dictionary method. The standard $k$SVD method cannot align the samples and the respective atoms only fit one pattern. The TWI-$k$SVD method manages to capture both patterns but the learned atoms yield unsmooth results. For the RISLDTW method, the learned mean series of aligned samples, as depicted in the left figure in Fig. \ref{dl_dictionary}(d), is affected by two patterns and represented a compromise between them. The right two figures show the eigenvectors corresponding to the first two largest eigenvalues, which are found to be uninterpretable. In contrast, our proposed GTWIDL method captures both patterns by two smooth atoms. This result suggests the superiority of our method in terms of temporal alignment and dictionary learning.

\label{res_dic}

\subsection{Evaluation on classification}
In this subsection, we compare the classification performance of our proposed GTWIDL algorithm with other benchmark methods mentioned in Section \ref{tunning}. The results are presented in Tab. \ref{Tab4} with the best performance highlighted in bold. It can be observed that our proposed GTWIDL method achieves the best overall performance, with a total number of 7 wins and an average ranking of 1.4.

From the results presented in Tab. \ref{Tab4}, we draw the following conclusions. Firstly, our proposed GTWIDL algorithm outperforms other state-of-the-art dictionary learning methods, such as TWI-kSVD, RISLDTW, and RISLDTW-ELS, on all datasets. This is consistent with the analysis presented in Section \ref{res_dic}, showing that our algorithm achieves better performance in both temporal alignment and dictionary learning. Secondly, the proposed GTWIDL method is also effective in dealing with multi-dimensional time series. Among the four handwritten character datasets, the GTWIDL algorithm achieved three wins and one runner-up. This demonstrates the superior design of the optimization problem given in (\ref{opt0}), where the sparseness coefficients and warping matrix are shared across all dimensions. Thirdly, the WDTW method also achieved good performance with wins on 6 out of 10 datasets. However, distance-based methods like WDTW cannot provide interpretable patterns like the proposed dictionary-based methods. Moreover, calculating each pair of samples using WDTW is time-consuming, which limits its applicability to large datasets. In contrast, our proposed GTWIDL algorithm offers interpretable and visible atoms to characterize patterns by learning dictionaries, which is beneficial for providing further analysis of each class's patterns. Moreover, our proposed optimization algorithm is time-efficient for large datasets since the number of atoms is much smaller than the number of samples.

\begin{table*}[tbh]
	%\large
	\caption{Classification accuracy comparison ({\textbf{Best}} in bold).}
	\label{Tab4}
	\centering
	%\resizebox{\columnwidth}{!}{
		\begin{tabular}{rccccccc}
			\hline
			& DTW             & DDTW            & WDTW            & TWI-kSVD        & { RISLDTW} & { RISLDTW-ELS} & ours            \\ \hline
			DIGITS              & 0.9852          & 0.9812          & 0.9860          & 0.9778          & 0.9802                         & 0.9788                             & \textbf{0.9976} \\
			LOWER               & 0.9459          & 0.9099          & 0.9473          & 0.9398          & 0.9345                         & 0.9411                             & \textbf{0.9888} \\
			UPPER               & 0.8589          & 0.8381          & 0.8630          & 0.8665          & 0.9139                         & 0.9101                             & \textbf{0.9885} \\
			char-traj           & 0.9398          & 0.8430          & \textbf{0.9605} & 0.9564          & 0.9416                         & 0.9420                             & 0.9595          \\
			UMD                 & 0.9889          & 0.9979          & \textbf{1.0000} & 0.9653          & 0.9951                         & \textbf{1.0000}                    & \textbf{1.0000}          \\
			BME                 & 0.9560          & 0.9893          & \textbf{1.0000} & \textbf{1.0000} & \textbf{1.0000}                & \textbf{1.0000}                    & \textbf{1.0000} \\
			ArrowHead           & 0.7029          & 0.7771          & 0.7486          & 0.7029          & 0.8286                         & 0.8286                             & \textbf{0.8514} \\
			DSR                 & \textbf{0.9673} & 0.9346          & \textbf{0.9673} & 0.9379          & 0.9248                         & 0.8399                             & 0.9608          \\
			FiftyWords          & 0.6967          & 0.6989          & \textbf{0.7714} & 0.7297          & 0.6484                         & 0.6659                             & 0.7429          \\
			Trace               & \textbf{1.0000} & \textbf{1.0000} & \textbf{1.0000} & 0.9900          & \textbf{1.0000}                & \textbf{1.0000}                    & \textbf{1.0000} \\ \hline
			Wins            & 2               & 1               & 6      & 1               & 2                              & 3                                  & \textbf{7}      \\
			Avg. rank           & 4.4             & 4.9             & 2.0             & 4.7             & 4.1                            & 3.5                                & \textbf{1.4}    \\
			\hline
		\end{tabular}
	%}
\end{table*}

\begin{table*}[tbh]
	%\large
	\caption{Clustering accuracy comparison ({\textbf{Best}} in bold).}
	\label{Tab5}
	\centering
	%\resizebox{\columnwidth}{!}{
		\begin{tabular}{rccccccc}
			\hline
			& SSC    & DTW             & DDTW            & WDTW            & TWI-kSVD & RISLDTW         & ours            \\
			DIGITS              & 0.8250 & 0.9390          & 0.9450          & 0.9290          & 0.9430   & 0.6580          & \textbf{0.9510} \\
			LOWER               & 0.6750 & 0.6769          & 0.6625          & 0.7202          & 0.6702   & 0.6327          & \textbf{0.7837} \\
			UPPER               & 0.6300 & 0.6796          & 0.7269          & 0.7385          & 0.6662   & 0.5638          & \textbf{0.7665} \\
			char-traj           & 0.6565 & 0.8010          & 0.6810          & \textbf{0.8110} & 0.7720   & 0.6240          & 0.8030          \\
			UMD                 & 0.4416 & 0.6061          & \textbf{0.6797} & 0.6580          & 0.5628   & 0.5758          & 0.6537          \\
			BME                 & 0.6474 & 0.6579          & 0.7947          & 0.6579          & 0.6579   & 0.6368          & \textbf{0.8158} \\
			ArrowHead           & 0.5833 & 0.5556          & \textbf{0.6944} & 0.5556          & 0.6111   & 0.5556          & \textbf{0.6944} \\
			DSR & 0.7500 & \textbf{0.9375} & 0.7500          & 0.8750          & 0.8750   & \textbf{0.9375} & \textbf{0.9375} \\
			FiftyWords          & 0.3156 & 0.3822          & 0.4044          & 0.3689          & 0.4156   & 0.3378          & \textbf{0.4267} \\
			Trace               & 0.5900 & 0.7500          & 0.9700          & 0.7500          & 0.8500   & 0.7800          & \textbf{1.0000} \\
			Wins                & 0      & 1               & 2               & 1               & 0        & 1               & \textbf{8}      \\
			Avg. rank           & 5.9    & 3.6             & 3.1             & 3.4             & 3.8      & 5.6             & \textbf{1.3}    
			\\
			\hline
		\end{tabular}
		%}
\end{table*}

\subsection{Evaluation on clustering}
In this subsection, we conducted a comparison between the proposed clustering algorithm and other benchmarks. The results are presented in Tab. \ref{Tab5} with the best performance highlighted in bold. Similar to the classification results, the proposed GTWIDL method achieved the best overall performance with a total number of best values (Wins) of 8 and an average ranking (Avg. rank) of 1.3.

Our results indicate that the traditional SSC method is inadequate in effectively clustering misaligned data and performs the worst. However, by using the SSC results as initialization, our GTWIDL method significantly improved the clustering accuracy by iteratively revising the clustering results. It is worth noting that the DDTW distance-based method also demonstrated good performance in some cases. This is likely because the DDTW distance utilizes derivatives as features, focusing on local patterns and thereby making it more robust for the clustering task.

\section{Conclusions}
\label{sec6}
This paper presents a generalized time warping invariant dictionary learning approach for capturing the underlying patterns of time series data subject to temporal warping. Our method can effectively handle multi-dimensional datasets with different sequence lengths by estimating the warping matrix as a linear combination of basis warping functions. Additionally, we introduce a block coordinate descent-based optimization algorithm that jointly solves for warping paths, dictionaries, and sparse coding coefficients, thereby improving time efficiency for large datasets. We then employ our proposed dictionary learning method to develop both classification and clustering algorithms, whereby samples are assigned to the class whose dictionary yields the minimum reconstruction error. Our experiments demonstrate that our proposed GTWIDL method provides interpretable atoms, which outperforms other dictionary learning methods in terms of representative and discriminative ability. Furthermore, our method outperforms other state-of-the-art classification and clustering benchmarks, which highlights its superior performance in various applications.

There are several interesting topics for further investigation based on this work. First, the current classification strategy only considers the reconstruction error on the local dictionary. Improved results may be achieved by proposing a classification strategy that takes both reconstruction error and sparseness coefficients into account. Second, since data can be aligned with sub-parts of atoms, the global data patterns could be recognized with few local observations at the beginning of a process. These patterns can be subsequently used for data prediction or system monitoring in the following data-collecting process. Finally, dictionary learning is widely used in image and video processing. The proposed one-directional time warping invariant dictionary learning has the potential to be extended to a generalized approach focusing on two-directional or three-directional spatio-temporal warping invariance.

%\section*{Acknowledgments}
%The authors would like to thank the editors and reviewers for providing thorough and thoughtful feedback which  leads to the substantial improvements of the paper.
%
%This research is supported by NSFC grant NSFC-72171003 and NSFC-71932006.

%\section{References Section}

\bibliographystyle{IEEEtran}
\bibliography{reference.bib}

\end{document}